\documentclass[10pt,twocolumn,letterpaper]{article}

\usepackage{iccv}
\usepackage{times}
\usepackage{epsfig}
\usepackage{graphicx}
\usepackage{amsmath}
\usepackage{amssymb}

\usepackage{graphicx}
\usepackage{booktabs} 
\usepackage{mathtools}
\usepackage{amsthm}
\usepackage{newfloat}
\usepackage{listings}
\usepackage{multirow}
\usepackage{bbm}
\usepackage{enumitem}
\usepackage[T1]{fontenc}
\usepackage[table]{xcolor}
\usepackage{subfigure}
\usepackage{cancel}
\usepackage{wrapfig}
\usepackage{appendix}
\usepackage{units}
\usepackage{xspace}
\usepackage{xcolor}
\usepackage{enumerate}
\usepackage{authblk}
\usepackage[ruled]{algorithm2e}
\usepackage[accsupp]{axessibility}

\theoremstyle{plain}
\newtheorem{theorem}{Theorem}[section]

\theoremstyle{definition}
\newtheorem{definition}[theorem]{Definition}

\theoremstyle{remark}

\newcommand{\h}{\mathbf{h}}

\newcommand{\FedAvg}{\textsc{FedAvg}\xspace}
\newcommand{\FedNH}{\textsc{FedNH}\xspace}
\newcommand{\FedProto}{\textsc{FedProto}\xspace}
\newcommand{\Ditto}{\textsc{Ditto}\xspace}
\newcommand{\FedRep}{\textsc{FedRep}\xspace}
\newcommand{\FedRoD}{\textsc{FedRoD}\xspace}
\newcommand{\FedETF}{\textsc{FedETF}\xspace}
\newcommand{\CCVR}{\textsc{CCVR}\xspace}
\newcommand{\FedDyn}{\textsc{FedDyn}\xspace}
\newcommand{\FedProx}{\textsc{FedProx}\xspace}

\def\bx{\mathbf{x}}

\def\bw{\mathbf{w}}

\def\bv{\mathbf{v}}
\def\DD{\mathcal{D}}

\def\bu{\mathbf{u}}
\def\bv{\mathbf{v}}

\def\bh{\mathbf{h}}
\def\bp{\mathbf{p}}
\def\bmu{\boldsymbol{\mu}}

\usepackage[breaklinks=true,bookmarks=false]{hyperref}
\usepackage{fontawesome5}

\iccvfinalcopy 

\def\iccvPaperID{****} 

\ificcvfinal\fi

\begin{document}

\title{No Fear of Classifier Biases: Neural Collapse Inspired Federated Learning with Synthetic and Fixed Classifier}

\author{Zexi Li$^1$\qquad Xinyi Shang$^{2\dag}$\qquad Rui He$^1$\qquad Tao Lin$^{3}$\thanks{Corresponding authors. $^\dag$Work was done during Xinyi's visit to Westlake University.}\qquad Chao Wu$^{1*}$\\
$^1$Zhejiang University\qquad $^2$Xiamen University\qquad $^3$Westlake University\\
{\tt\small \{zexi.li,ruihe,chao.wu\}@zju.edu.cn}\quad {\tt\small shangxinyi@stu.xmu.edu.cn}\quad {\tt\small lintao@westlake.edu.cn}

}

\maketitle
\ificcvfinal\thispagestyle{empty}\fi

\begin{abstract}
Data heterogeneity is an inherent challenge that hinders the performance of federated learning (FL). Recent studies have identified the biased classifiers of local models as the key bottleneck. Previous attempts have used classifier calibration after FL training, but this approach falls short in improving the poor feature representations caused by training-time classifier biases. Resolving the classifier bias dilemma in FL requires a full understanding of the mechanisms behind the classifier. Recent advances in neural collapse have shown that the classifiers and feature prototypes under perfect training scenarios collapse into an optimal structure called simplex equiangular tight frame (ETF). Building on this neural collapse insight, we propose a solution to the FL's classifier bias problem by utilizing a synthetic and fixed ETF classifier during training. The optimal classifier structure enables all clients to learn unified and optimal feature representations even under extremely heterogeneous data. We devise several effective modules to better adapt the ETF structure in FL, achieving both high generalization and personalization. Extensive experiments demonstrate that our method achieves state-of-the-art performances on CIFAR-10, CIFAR-100, and Tiny-ImageNet. The code is available at \color{magenta}\url{https://github.com/ZexiLee/ICCV-2023-FedETF}\color{black}.
\end{abstract}

\section{Introduction}
Federated learning (FL) \cite{mcmahan2017communication,li2022towards,zhang2022dense,10.1145/3539618.3591976} is a distributed training paradigm that enables collaborative training from massive multi-source datasets without transferring the raw data, reserving data ownership \cite{li2022can} while relieving communication burdens \cite{mcmahan2017communication}. FL facilitates broad applications in medical images \cite{adnan2022federated,guo2022auto}, the internet of things \cite{khan2021federated,nguyen2021federated}, mobile services \cite{hard2020training,kang2020reliable}, and so on; it shows promising prospects in data collaboration. 
However, clients in FL training may hold heterogeneous data, in other words, clients' datasets are in Non-IID distributions\footnote{We use ``data heterogeneity'' and ``Non-IID data'' interchangeably.}, which causes a huge degradation to the global model's generalization \cite{dai2022tackling,luo2021no,chen2021bridging,li2023revisiting,FCCL_CVPR22,FPL_CVPR23}. 

\begin{figure*}[t]
\label{fig:motivation}
\centering
\includegraphics[width=1.83\columnwidth]{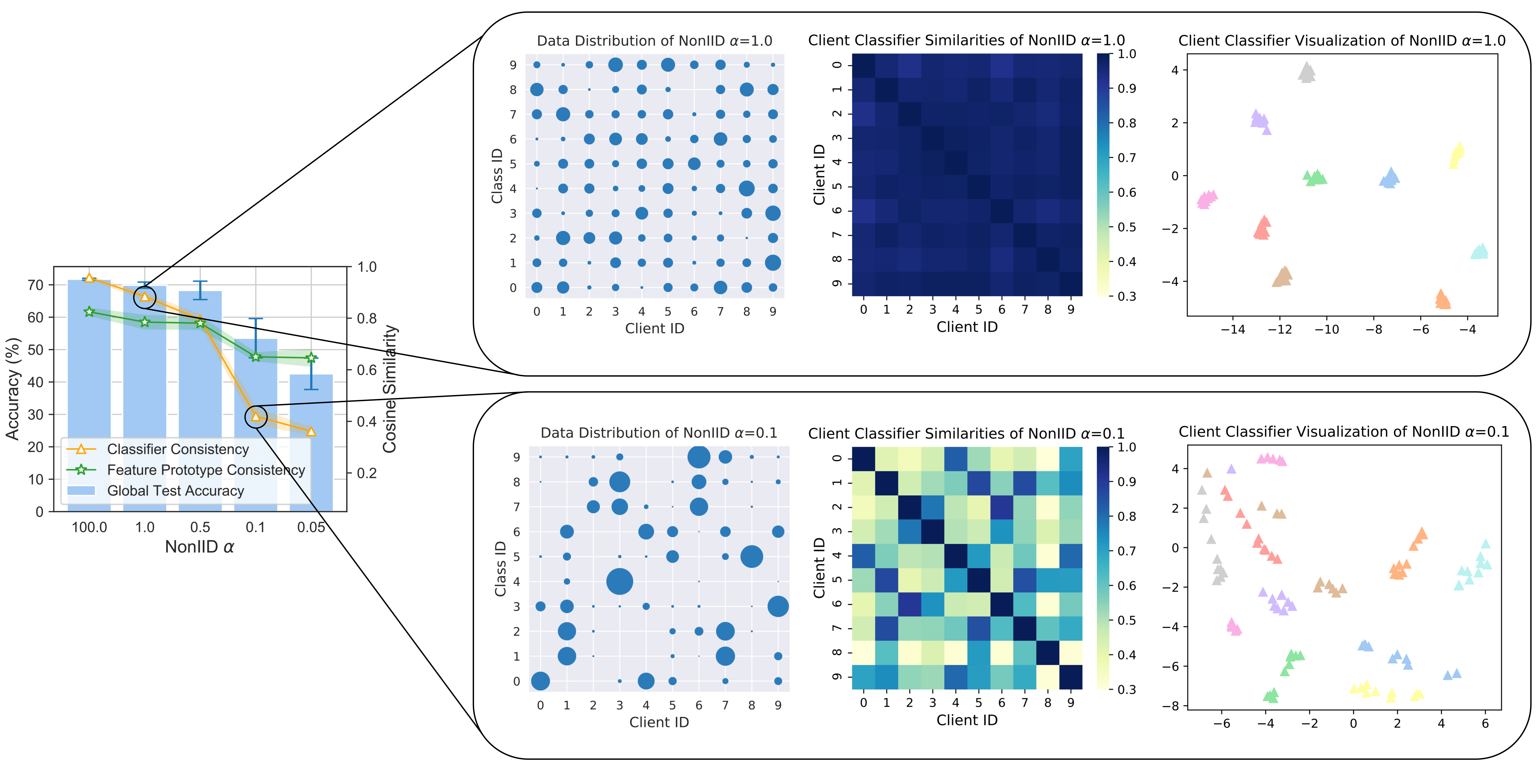}
\caption{ \textbf{How data heterogeneity causes classifier biases in FL.} Smaller $\alpha$ corresponds to higher Non-IID. Experiments are conducted on CIFAR-10 with vanilla \textsc{FedAvg}. Columns from left to right: (1) Non-IID data results in poor generalization, biased classifiers, and misaligned features. (2) Clients' data distributions. (3) Clients with Non-IID data have smaller pair-wise classifier cosine similarities. (4) t-SNE visualization of clients' class-wise classifier vectors (represented by colors), which are more scattered in Non-IID data. }
\end{figure*}

Numerous recent studies have shown that \textit{\textbf{classifier biases}} in clients' local models caused by Non-IID data are the primary cause of degradation in FL \cite{luo2021no,zhou2022fedfa,li2022partial}. It has been discovered that the classifier layer is more biased than other layers \cite{luo2021no}, and classifier biases will create a \textit{vicious cycle} between biased classifiers and misaligned features across clients \cite{zhou2022fedfa}. Figure \ref{fig:motivation} illustrates the issue of classifier bias in FL, where Non-IID data leads to poor pairwise cosine similarities among clients' classifiers and feature prototypes. Furthermore, class-wise classifier vectors of clients are scattered in the embedding space, leading to significant generalization declines.

Previous research has attempted to mitigate classifier biases through classifier retraining via generated virtual features at the end of FL training \cite{luo2021no,shang2022federated}. However, these methods fail to address classifier biases during training. Biased classifiers during the training phase lead to inadequate feature extractors and poor representations of generated features, negatively affecting the retrained classifiers. Our experiments have also shown the limitations of classifier retraining methods (Table \ref{table:first_table}). Therefore, we wonder:

\vspace{0.15cm}
\noindent\emph{Can we break the classifier bias dilemma during training, improving both the classifiers and feature representations?}
\vspace{0.15cm}

To resolve the classifier bias dilemma, it is essential to fully understand the mechanisms behind the classifier. We further wonder: \textit{what are the properties of a well-trained (i.e. good) classifier?} An emerging discovery called neural collapse \cite{papyan2020prevalence,yang2022we,li2022principled,yangneural} has shed light on this matter. It describes the phenomenon that, in the perfect training scenario, where the dataset is balanced and sufficient, the feature prototypes and classifier vectors converge to an optimal simplex equiangular tight frame (ETF) with maximal pairwise angles \cite{papyan2020prevalence}. Insights from balanced training inspire us to tackle the challenges in Non-IID FL.

Thus, in this paper, \textit{we \textbf{fundamentally solve the FL's classifier bias problem} with a neural-collapse-inspired approach. }
Knowing the optimal classifier structure, we make the first attempt to introduce a synthetic simplex ETF as a fixed classifier for all clients so that the clients can learn unified and optimal feature representations even under high heterogeneity. We devise \textsc{FedETF} which incorporates several effective modules that better adapt the ETF structure in FL training, reaching strong results on \textit{\textbf{both}} \textit{\textbf{generalization}} \textbf{\textit{and}} \textit{\textbf{personalization}}. 

Specifically, we employ a projection layer that maps the features to a space where neural collapse is more likely to occur. We also implement a balanced feature loss with a learnable temperature to minimize entropy between the features and the fixed ETF classifier. These techniques enable us to achieve high generalization performance of the global model during FL training. 
To further improve personalization, we introduce a novel fine-tuning strategy that adapts the global model locally after FL training. 
Extensive experiments have strongly supported the effectiveness of our method. Our contributions are summarized as follows.
\begin{itemize}[leftmargin=*,nosep]
    \item To the best of our knowledge, this is the first paper that tackles the data heterogeneity problem in FL from the perspective of neural collapse.
    \item We devise \textsc{FedETF}, which takes the simplex ETF as a fixed classifier, and it fundamentally solves the classifier biases brought by Non-IID data, reaching high generalization of the global model.
    \item We propose a local fine-tuning strategy in \FedETF to boost personalization in each client after FL training.
    \item Our method is validated on three vision datasets: CIFAR-10, CIFAR-100, and Tiny-ImageNet. Our proposed method outperforms strong baselines and achieves sota in both generalization and personalization.
\end{itemize}

\section{Related Works}
\noindent \textbf{Data Heterogeneity in Federated Learning.}
A variety of solutions have been proposed to tackle data heterogeneity in FL. Recent works \cite{li2022partial,luo2021no,shang2022federated,zhou2022fedfa} have revealed that the biased classifier is the main cause leading to poor performance of the global model, and they use classifier retraining \cite{luo2021no,shang2022federated} or classifier variance reduction \cite{li2022partial} to calibrate the classifier. In particular, \CCVR \cite{luo2021no} finds that there exists a greater bias in the classifier than in other layers, and only calibrating the classifier via virtual features after FL training can improve the global model performance. However, this approach cannot resolve the misaligned representations of local models caused by biased classifiers during FL training. Consequently, the backbone feature extractor cannot be improved. Moreover, some concurrent works use classifier variance reduction \cite{li2022partial} or feature anchors \cite{zhou2022fedfa} to relieve classifier biases. However, when data is highly Non-IID, variance-reduced classifiers and aggregated feature anchors are also biased and far from optimal (i.e., trained on IID data). Although these methods can alleviate classifier biases to some extent, they cannot completely solve them. 

To improve the misaligned features, another line of works use prototypical methods to aid client training. \FedProto \cite{tan2022fedproto} only transmits and aggregates prototypes on the server to deal with data heterogeneity and model heterogeneity, and \textsc{FedFM} anchors clients' features to improve generalization \cite{ye2022fedfm}. A concurrent work named \FedNH \cite{dai2022tackling} adopts a prototypical classifier and uses a smoothing aggregation strategy to update the classifier based on clients' local prototypes. 
However, the aggregated prototypes will also be biased in extreme Non-IID settings, and these methods did not use the classifier's ETF optimality to tackle this problem, which is where our contribution lies. 

Besides, data heterogeneity can be relieved by improving aggregation \cite{li2023revisiting,ye2023feddisco}. Apart from generalization, there are data heterogeneity challenges for personalization, methods such as decoupling classifiers and feature extractors \cite{chen2021bridging}, separating feature information \cite{fedcp}, adaptive local aggregation \cite{zhang2023fedala}, and edge-cloud collaboration \cite{10.1145/3539618.3591976} can be used to facilitate better local personalized models.

\noindent \textbf{Neural Collapse.} 
The neural collapse was firstly observed in \cite{papyan2020prevalence} that at the terminal phase of training on a balanced dataset, the feature prototypes and the classifier vectors will converge to a simplex ETF where the vectors are normalized and the pair-wise angles are maximized. Afterward, there are some works trying to figure out the mechanism behind neural collapse \cite{ji2021unconstrained,zhu2021geometric,tirer2022extended,kothapalli2022neural} and in which conditions neural collapse will happen \cite{li2022principled,tirer2022extended,kothapalli2022neural}. Recent works use neural-collapse-inspired methods to solve the problems in imbalanced training \cite{yang2022we,xie2023neural,thrampoulidis2022imbalance}, incremental learning \cite{yangneural}, and transfer learning \cite{li2022principled}. 
Despite the neural-collapse-inspired methods' success in centralized learning, deep insights and effective solutions regarding neural collapse are missing in distributed training. In this paper, we find neural collapse is also the key to success in FL and we show that inducing ETF optimality can inherently solve the classifier bias and feature misalignment problem in FL and largely improve performance. 
\section{Preliminaries}
\subsection{Federated Learning}
\noindent\textbf{Basic Settings.} We introduce a typical FL setting with $K$ clients holding potentially Non-IID data partitions $\DD_1, \DD_2, ..., \DD_K$, respectively. A supervised classification task with $C$ classes is considered. Let $n_{k,c}$ be the number of samples of class $c$ on client $k$, and $n_k=\sum_{c=1}^Cn_{k,c}=|\DD_k|$ denotes the number of training samples held by client $k$. FL aims to realize generalization or personalization by distributed training under the coordination of a central server without any data transmission. 
For \textit{generalization} in FL, the goal is to learn a global model over the whole training data $\DD\triangleq \bigcup_{k} \DD_k$ and the global model is expected to be generalized to the distribution of the whole data $\DD$. For \textit{personalization}, the goal is to learn $K$ personalized local models by FL training, and the local model $k$ is expected to have better adaptation in local data distribution $\DD_k$ than the independently trained one. 
For the model in FL, we typically consider a neural network $\phi_{\bw}$ with parameters $\bw=\{\bu, \bv\}$. It has two main components: 1) a feature extractor $f_{\bu}$ parameterized by $\bu$, mapping each input sample $\bx$ to a $d$-dim feature vector; 2) a classifier $h_{\bv}$ parameterized by $\bv$. The parameters of client $k$'s local model are denoted as $\bw_k$.

\noindent\textbf{Model updates.} There are two iterative steps in FL, client local training and server global aggregation, and the iteration lasts for $T$ rounds. In round $t$, the server first sends a global model $\bw^t$ to clients. 

\noindent\textit{\textbf{Client local training:}} For each client $k,~ \forall k \in [K]$, it conducts SGD updates on the local data $\DD_k$:
\begin{align}\label{equ:client_update}
\bw_k^{t}\gets \bw_k^t-\eta\nabla_{\bw}\ell(\bw_k^t;\mathcal{B}^i),
\end{align}
where $\eta$ is the learning rate, $\mathcal{B}^i$ is the mini-batch sampled from $\DD_k$ at the $i$-th local iteration. The local epoch is $E$. 

\noindent\textit{\textbf{Server global aggregation:}} 
After local training, the server samples a set of clients $\mathcal{A}^t$ and the sampled clients send their updated models to the server. Then the server performs the weighted aggregation to update the global model for round $t+1$. The vanilla aggregation strategy is \FedAvg \cite{mcmahan2017communication} where the aggregation weights are proportional to clients' data sizes. 
\begin{align}\label{equ:global_agg}
\bw^{t+1}&= \sum_{k\in \mathcal{A}^t}\frac{n_k}{\sum_{j\in\mathcal{A}^{t}}n_j}\bw_k^{t}.
\end{align}

\subsection{Neural Collapse} \label{sect:pre_nc}
Neural collapse refers to a phenomenon about the last-layer features and classifier vectors at the terminal phase of training (zero training loss) on a balanced dataset \cite{papyan2020prevalence}. We first give the definition of simplex ETF in neural collapse. 
\begin{definition}[Simplex Equiangular Tight Frame]
	\label{def:ETF}
A collection of vectors $\mathbf{v}_i\in\mathbb{R}^d$, $i \in [C]$, $d\ge C-1$, is said to be a simplex equiangular tight frame if:
\begin{equation}\label{equ:ETF_definition}
    \mathbf{V}=\sqrt{\frac{C}{C-1}}\mathbf{U}\left(\mathbf{I}_C-\frac{1}{C}\mathbf{1}_C\mathbf{1}_C^T\right),
\end{equation}
where $\mathbf{V}=[\mathbf{v}_1,\cdots,\mathbf{v}_C]\in\mathbb{R}^{d\times C}$, $\mathbf{U}\in\mathbb{R}^{d\times C}$ allows a rotation and satisfies $\mathbf{U}^T\mathbf{U}=\mathbf{I}_C$, $\mathbf{I}_C$ is the identity matrix, and $\mathbf{1}_C$ is an all-ones vector. All vectors in a simplex ETF have an equal $\ell_2$ norm and the same pair-wise angle, i.e.
\begin{equation}\label{equ:mimj}
	\mathbf{v}_i^T\mathbf{v}_j=\frac{C}{C-1}\delta_{i,j}-\frac{1}{C-1}, \forall i, j\in[C],
\end{equation}
where $\delta_{i,j}$ equals to $1$ when $i=j$ and $0$ otherwise. The pair-wise angle $-\frac{1}{C-1}$ is the maximal equiangular separation of $C$ vectors in $\mathbb{R}^d$ \cite{papyan2020prevalence}.
\end{definition}

We highlight three key properties of the neural collapse (NC) phenomenon below.

\noindent\textbf{NC1} (Features collapse to the class prototypes). 
The last-layer features will collapse to their within-class mean (prototypes), i.e. for any class $c,~ \forall c \in [C]$, the covariance $\Sigma_W^c\rightarrow\mathbf{0}$, where $\Sigma_W^c:=\frac{1}{n_c}\sum_{i=1}^{n_c}(\h_{c,i}-\h_c)(\h_{c,i}-\h_c)^T$. $\h_{c,i}=f(\bu;\bx_{c,i})$ is the feature of the $i$-th sample in the class $c$, and $\h_c=\frac{1}{n_c}\sum_{i=1}^{n_c}\h_{c,i}$ is the class $c$'s prototype.

\noindent\textbf{NC2} (Prototypes collapse to simplex ETFs).  $\tilde{\h}_c = (\h_c-\h_G)/||\h_c-\h_G||, \forall c\in[C]$, collapses to a simplex ETF which satisfies Eq. (\ref{equ:mimj}). $\h_G$ is the global mean of the last-layer features, that $\h_G=\sum_{c=1}^C\sum_{i=1}^{n_c}\h_{c,i}$.

\noindent\textbf{NC3} (Classifiers collapse to the same simplex ETFs). The normalized feature prototype $\tilde{\h}_c$ is aligned with their corresponding classifier weights\footnote{For simplicity, we omit the bias term in a linear classifier layer.}, which means that the classifier weights collapse to the same simplex ETF, i.e. $\tilde{\h}_c=\bv_c/||\bv_c||$, where $\bv_c$ refers to the vectorized classifier weights of class $c$.

\begin{figure*}[t]
    \centering
    \includegraphics[width=2.2\columnwidth]{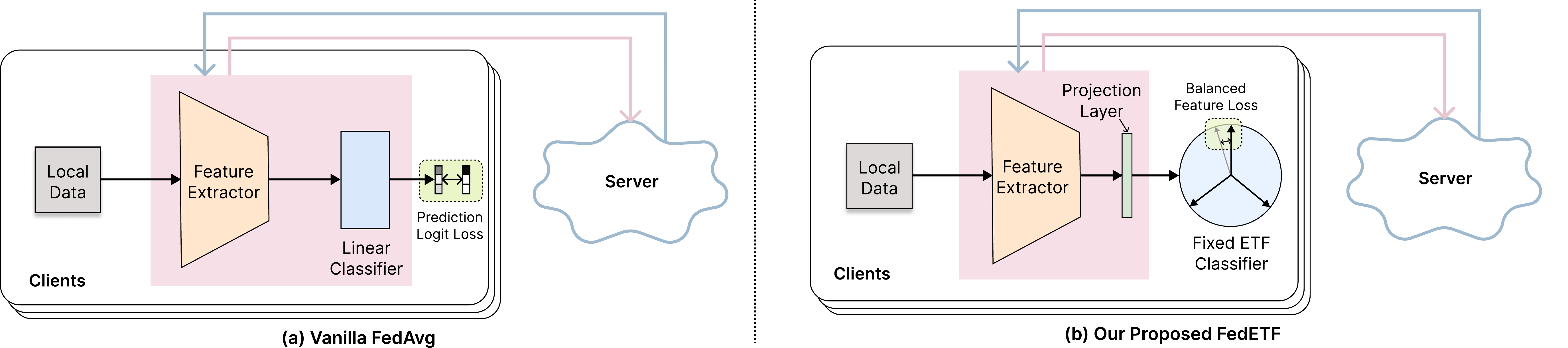}
    \caption{\textbf{Proposed \FedETF during FL training.} (a) In vanilla FL training, the feature extractor and linear classifier are both learned at clients and aggregated at server. (b) In our \FedETF, only the feature extractor and projection layer are learned and aggregated, and we adopt the same synthetic and fixed ETF classifier for all clients throughout the FL training process. Instead of prediction logit loss in vanilla FL, we use a novel balanced feature loss for the ETF classifier.}
    \label{fig:framework}
\end{figure*}

\section{Methods}
Neural collapse tells us the optimal structure (i.e. simplex ETF) of classifiers and feature prototypes in a perfect training setting. It inspires us to use a synthetic simplex ETF as a fixed classifier from the start to mitigate the classifier bias and feature misalignment problems (see Figure~\ref{fig:motivation}) brought by clients' data heterogeneity. Therefore, we propose \FedETF, a novel FL algorithm inspired by neural collapse. Concretely, as elaborated in Section \ref{sect:fedetf_generalization}, to promote the generalization of the global model, we reformulate the model architecture by replacing the learnable classifier with a fixed ETF classifier and devise a tailored loss for robust learning during FL training. Moreover, as described in Section \ref{sect:fedetf_personalization}, to improve local personalization after FL training, we propose a finetuning strategy for both finetuning the model and the formerly fixed ETF classifier. 

\subsection{Improving Generalization by ETF Classifier}
\label{sect:fedetf_generalization}
\noindent\textbf{Reformulation of the model architecture.} In previous works of FL, a model architecture that consists of a learnable feature extractor and a learnable linear classifier is adopted \cite{mcmahan2017communication,chen2021bridging,collins2021exploiting,li2020federated}, as shown in Figure \ref{fig:framework} (a). However, due to clients' data heterogeneity, the classifiers will be more biased than other layers \cite{luo2021no,li2022partial}, and a vicious cycle between classifier biases and feature misalignment will exist \cite{zhou2022fedfa}. In this paper, we reformulate the model in FL into the combination of a learnable feature extractor and a fixed ETF classifier, as demonstrated in Figure \ref{fig:framework} (b).

\noindent\textit{\textbf{ETF classifier initialization:}} At the beginning of the FL training, we first randomly synthesize a simplex ETF $\mathbf{V}_{ETF} \in \mathbb{R}^{d\times C}$ by Eq. (\ref{equ:ETF_definition}), where $d$ denotes the feature dimension of the ETF and $C$ is the number of classes. The feature dimension $d$ should require $d\ge C-1$ in Definition \ref{def:ETF}, and we will discuss in Figure \ref{fig:understanding_fedetf_featuredim_personalization} (a) that a relatively low dimension is beneficial to neural collapse. For each class's classifier vector $\bv_i,~ \forall i\in [C]$ in the ETF $\mathbf{V}_{ETF}$, it requires $\Vert\bv_i \Vert_2 = 1$; and any pair of classifier vectors $(\bv_i,\bv_j),~ i \neq j,~ \forall i,j\in [C]$ satisfies $\cos(\bv_i,\bv_j) = -\frac{1}{C-1}$ according to Eq. (\ref{equ:mimj}). 

\noindent\textit{\textbf{Projection layer:}} Given a data sample $\bx$, we first use the feature extractor $f_\bu$ to transform the data into the raw feature $\mathbf{h}$ and then use a projection layer $g_{\mathbf{p}}$ to map this raw feature to the ETF feature space and normalize it into $\bmu$. 
\begin{equation} \label{equ:etf_feature}
    \bmu = \hat{\bmu} / \Vert \hat{\bmu} \Vert_2,\quad \hat{\bmu} = g(\bp;\bh),\quad \bh = f(\bu;\bx), 
\end{equation}
where $\bu$ and $\bp$ denote the parameters of the feature extractor and the projection layer. We note that the projection layer is essential in our \FedETF design: 1) If the last layer of the feature extractor is the non-linear activation, e.g. the ReLU, the raw feature $\bh$ will be sparse with zeros (or near zero values), and it is hard for $\bh$ to be close to the dense ETF classifier vectors. 2) The raw features always have high dimensions, and high-dimensional vectors are more prone to be orthogonal, which is harder to collapse into the ETF with maximal angles. It is necessary to use the projection layer to map the features into a suitable dimension $d$. 3) The projection layer is helpful in the local finetuning stage for personalization.

\noindent\textbf{Balanced feature loss with learnable temperature.} In neural-collapse-inspired imbalanced learning \cite{yuan2021we}, it is found that when the ETF classifier is used, the gradients of cross entropy (CE) will be biased towards the head class, and the authors proposed a dot regression loss to tackle this problem. In FL, clients' local datasets are also class-imbalanced due to data heterogeneity, so techniques tackling the imbalanced problem are also needed in our design. Following previous work \cite{chen2021bridging} which induces balanced softmax loss to logit-prediction-based CE in FL, in this paper, we also incorporate the balanced loss \cite{ren2020balanced} into our feature-based CE. Instead of using former dot regression loss \cite{yuan2021we,yangneural}, it is found that our balanced feature CE loss also can solve the imbalanced gradient problem in learning with the ETF classifier. 

Moreover, the softmax function's input of the vanilla CE loss is the logits, generated by MLP, while that of our method is the features' product $\bv_{y}^T\bmu$. The logits have a wide range of value since the output of MLP has no constraints, but our features' product has a limited range [-1, 1]\footnote{Knowing the fact that both the vectors are normalized.}, which is sensitive to scaling. Therefore, we add a temperature\footnote{The term ``temperature'' is borrowed from a similar concept in knowledge distillation \cite{hinton2015distilling}.} scalar $\beta$ to scale the features' product. Further, we found that in different stages of training, it requires different $\beta$, and fixed $\beta$ is hard to tune and may impede performance if not appropriate. To solve this, we take $\beta$ as one of the parameters in the model and update it by SGD during training. This learnable temperature $\beta$ will capture the learning dynamics in each client under various heterogeneity.

We define the model parameters in our \FedETF as $\bw=\{\bu, \bp, \beta\}$, which consists of the feature extractor, the projection layer, and the learnable temperature. For a given sample $(\bx, y)$ of client $k,~ \forall k \in [K]$, we define the loss function for generalization in Eq. (\ref{equ:g_sample_loss}), where the \textcolor{orange}{orange} term is for balanced feature loss and the \textcolor{blue}{blue} term is for learnable temperature. 
\begin{align} \label{equ:g_sample_loss}
\small
    \ell^g(\bw, \mathbf{V}_{ETF}; \bx, y) = -\log\frac{\textcolor{orange}{n_{k,y}^{\gamma}}\exp(\textcolor{blue}{\beta}\cdot\bv_{y}^T\bmu)}{\sum_{c\in[C]}\textcolor{orange}{n_{k,c}^{\gamma}}\exp(\textcolor{blue}{\beta}\cdot\bv_{c}^T\bmu)},
\end{align}
where $n_{k,c}$ refers to the number of samples in class $c$ of client $k$, $\beta$ is the learnable temperature, $\bmu$ is the normalized feature in Eq. (\ref{equ:etf_feature}), $\bv_{c}$ is the class $c$'s classifier vector in $\mathbf{V}_{ETF}$, and $\gamma$ is the hyperparameter for balanced loss. Below, we give the learning objective of client $k,~ \forall k \in [K]$, and solve the objective by SGD in Eq. (\ref{equ:client_update}).
\begin{align}
    \bw_k^t &= \arg\min_{\bw} \mathcal{L}_k^g(\bw), \\
    \text{where }\mathcal{L}_k^g(\bw) &= \frac{1}{n_k}\sum_{(\bx_i, y_i)\in \DD_k}\ell^g(\bw, \mathbf{V}_{ETF}; \bx_i, y_i).
\end{align}
We adopt the vanilla aggregation strategy on the server, formulated in Eq. (\ref{equ:global_agg}). The pseudo code of the proposed \textsc{FedETF} is shown in Algorithm 1 in Appendix.

\begin{figure}[t]
    \centering
    \includegraphics[width=0.9\columnwidth]{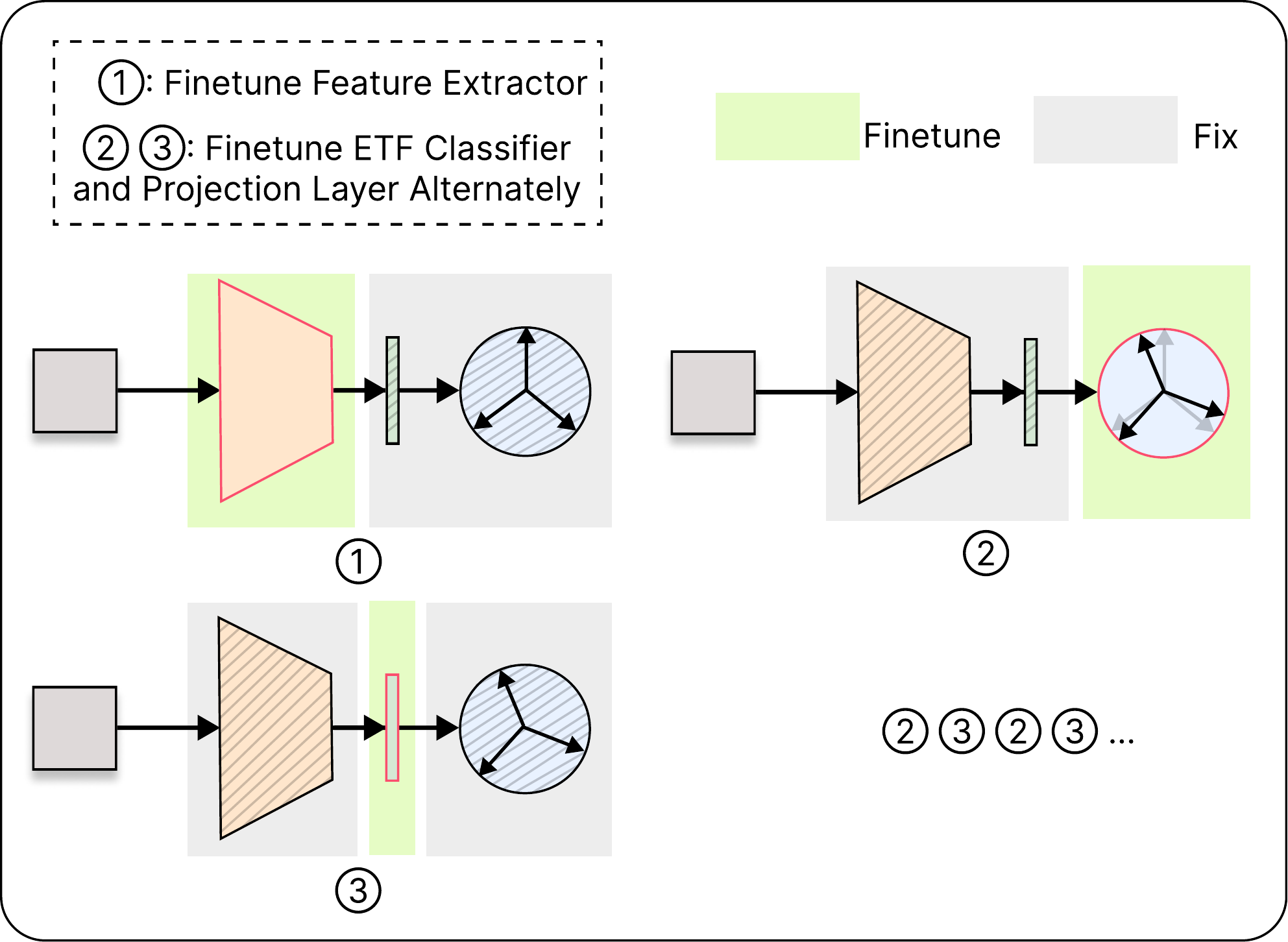}
    \caption{\textbf{Local finetuning stage of proposed \FedETF for personalization.} This stage is after the whole FL training stage when clients receive the final global model. For each client, we first finetune the feature extractor and then we finetune the ETF prototypical classifier and projection layer alternately.}
    \label{fig:fedetf_personalization_framework}
\end{figure}

\begin{table*}[!ht]
    \footnotesize
    \centering
    \caption{\textbf{Results in terms of generalization (General.) accuracy (\%) of global models and personalization (Personal.) accuracy (\%) of local models on three datasets under different heterogeneity.} Best two methods in each setting are highlighted in \textbf{bold} fonts.}
    \resizebox{\linewidth}{!}{
    \begin{tabular}{l|cc|cc|cc|cc|cc|cc}
    \toprule
    Dataset&\multicolumn{4}{c}{CIFAR-10}&\multicolumn{4}{c}{CIFAR-100}&\multicolumn{4}{c}{Tiny-ImageNet}\\
    \cmidrule(lr){1-5}
    \cmidrule(lr){6-9}
    \cmidrule(lr){10-13}
    NonIID ($\alpha$) &\multicolumn{2}{c}{0.1}&\multicolumn{2}{c}{0.05}&\multicolumn{2}{c}{0.1}&\multicolumn{2}{c}{0.05}&\multicolumn{2}{c}{0.1}&\multicolumn{2}{c}{0.05}\\
    \midrule
    Methods/Metrics &General.&Personal.&General.&Personal.&General.&Personal.&General.&Personal.&General.&Personal.&General.&Personal.\\
    \midrule
    \textsc{FedAvg} \cite{mcmahan2017communication}&52.76{\tiny±6.08}	&83.85{\tiny±0.89}	
    &44.48{\tiny±6.19}	&89.80{\tiny±0.39}
    &24.77{\tiny±1.19}	&49.93{\tiny±1.17}	&\textbf{22.53{\tiny±0.40}}	&58.85{\tiny±0.33}	&28.93{\tiny±0.52}	&40.81{\tiny±0.35}	&24.88{\tiny±0.34}	&46.90{\tiny±0.44}\\
    \midrule
    \textsc{FedProx} \cite{li2020federated}&46.59{\tiny±3.04}	&82.08{\tiny±0.27}	
    &40.95{\tiny±5.75}	&87.69{\tiny±2.85}
    &23.33{\tiny±1.72}	&46.44{\tiny±1.64}	&19.12{\tiny±0.77}	&57.01{\tiny±2.17}	&25.93{\tiny±0.27}	&31.90{\tiny±1.91}	&23.06{\tiny±0.68}	&32.43{\tiny±0.65}\\
    \textsc{FedDyn} \cite{acar2020federated}&36.35{\tiny±5.33}	&85.39{\tiny±0.77}	
    &23.90{\tiny±1.40}	&88.72{\tiny±1.59}
    &\textbf{25.53{\tiny±2.39}}	&51.79{\tiny±2.12}	&20.71{\tiny±2.83}	&\textbf{61.77{\tiny±0.32}}	&26.42{\tiny±0.56}	&45.84{\tiny±0.34}	&23.63{\tiny±1.55}	&52.27{\tiny±1.06}\\
    \midrule
    \textsc{Ditto} \cite{li2021ditto}&52.76{\tiny±6.08}	&79.81{\tiny±1.89}	
    &44.48{\tiny±6.19}	&85.17{\tiny±3.47}
    &24.77{\tiny±1.19}	&38.06{\tiny±1.26}	&22.53{\tiny±0.40}	&50.18{\tiny±1.22}	&28.93{\tiny±0.52}	&33.00{\tiny±1.01}	&24.88{\tiny±0.34}	&40.31{\tiny±0.12}\\
    \textsc{FedRep} \cite{collins2021exploiting}&26.85{\tiny±10.13}	&\textbf{87.76{\tiny±0.87}}	
    &15.79{\tiny±3.68}	&90.71{\tiny±2.25}
    &5.47{\tiny±0.20} &\textbf{53.62{\tiny±1.49}}	&4.18{\tiny±0.85}	&61.51{\tiny±0.61}	&4.10{\tiny±0.22}	&43.66{\tiny±0.48}	&2.20{\tiny±0.19}	&49.52{\tiny±1.64}\\
    \midrule
    \textsc{CCVR} \cite{luo2021no}&52.50{\tiny±6.31}	&55.62{\tiny±5.89}	&47.98{\tiny±6.24}	&73.52{\tiny±7.49}	&24.54{\tiny±0.71}	&34.01{\tiny±2.01}	&22.28{\tiny±0.43}	&39.16{\tiny±1.41}	&\textbf{32.78{\tiny±0.24}}	&\textbf{54.00{\tiny±0.46}}	&\textbf{29.27{\tiny±0.25}}	&\textbf{59.29{\tiny±0.30}}	\\
    \textsc{FedProto} \cite{tan2022fedproto}&-	&83.34{\tiny±0.71}	
    &-	&88.21{\tiny±1.77}
    &- &43.31{\tiny±0.70}	
    &- &54.87{\tiny±0.52}	
    &-	&40.74{\tiny±0.87}	&-	&48.05{\tiny±0.82}\\
    \textsc{FedRoD} \cite{chen2021bridging}&\textbf{55.72{\tiny±2.40}}	&86.19{\tiny±0.91}	
    &\textbf{49.89{\tiny±3.64}}	&88.83{\tiny±4.14}
    &24.49{\tiny±1.05} &51.78{\tiny±1.16}	&21.63{\tiny±0.42}	&59.44{\tiny±0.45}	&32.17{\tiny±0.41}	&38.27{\tiny±1.00}	&28.45{\tiny±0.58}	&44.09{\tiny±0.44}\\
    \textsc{FedNH} \cite{dai2022tackling}&55.37{\tiny±4.48}	&85.98{\tiny±0.15}	
    &47.96{\tiny±2.59}	&\textbf{91.06{\tiny±3.13}}
    &24.67{\tiny±0.68} &52.09{\tiny±0.78}	&21.95{\tiny±0.85}	&\textbf{62.71{\tiny±0.22}}	&17.51{\tiny±0.62}	&36.53{\tiny±0.29}	&14.00{\tiny±0.17}	&41.80{\tiny±1.78}\\
    \midrule
    \rowcolor{gray!20}\textbf{Our \textsc{FedETF}}&\textbf{59.56{\tiny±1.84}}	&\textbf{87.89{\tiny±1.19}}	&\textbf{56.08{\tiny±3.44}}	&\textbf{92.62{\tiny±0.54}}
    &\textbf{26.24{\tiny±1.78}}	&\textbf{52.86{\tiny±1.53}}	&\textbf{24.17{\tiny±0.54}}	
    &60.68{\tiny±0.91}	
    &\textbf{33.49{\tiny±0.82}}	&\textbf{55.82{\tiny±0.60}}	&\textbf{29.15{\tiny±1.03}}	&\textbf{62.36{\tiny±0.13}}\\
    \bottomrule
    \end{tabular}
    }
    \label{table:first_table}
\end{table*}

\begin{figure*}[t]
\centering
\subfigure[]{\includegraphics[width=0.5\columnwidth]{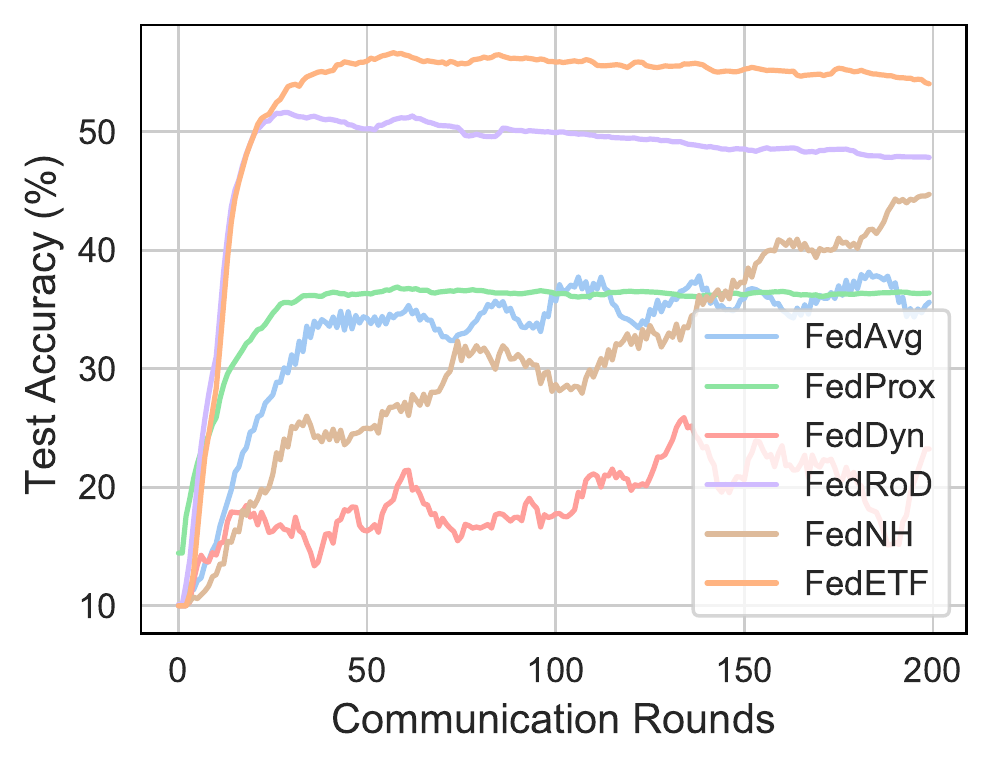}}
\subfigure[]{\includegraphics[width=0.5\columnwidth]{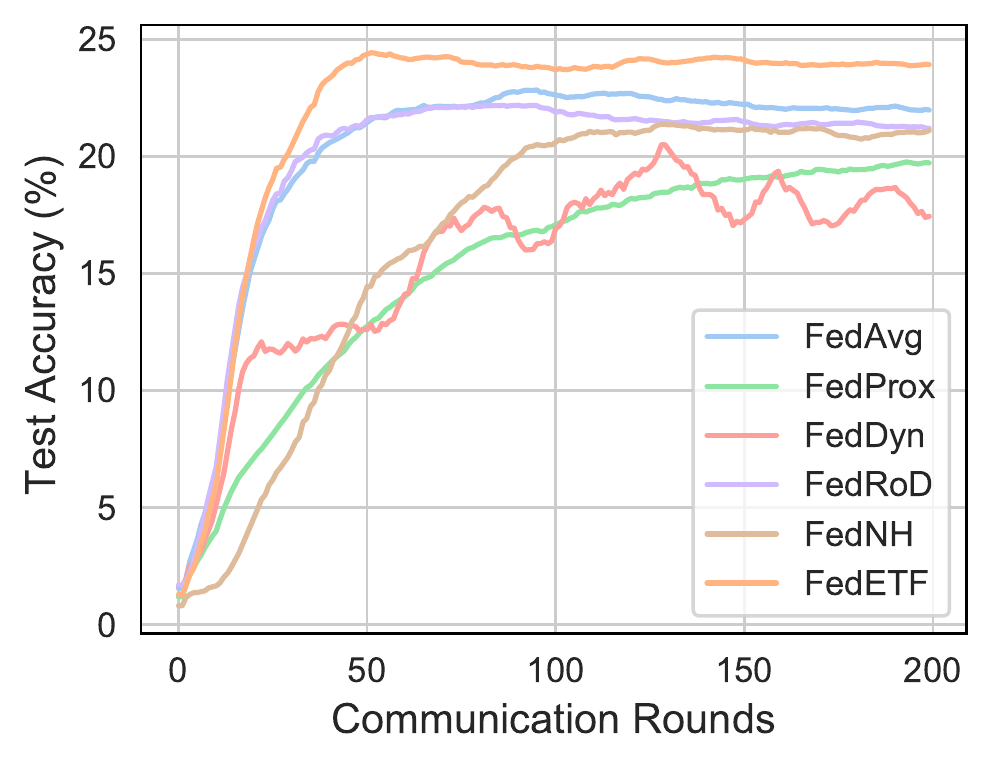}}
\subfigure[]{\includegraphics[width=0.5\columnwidth]{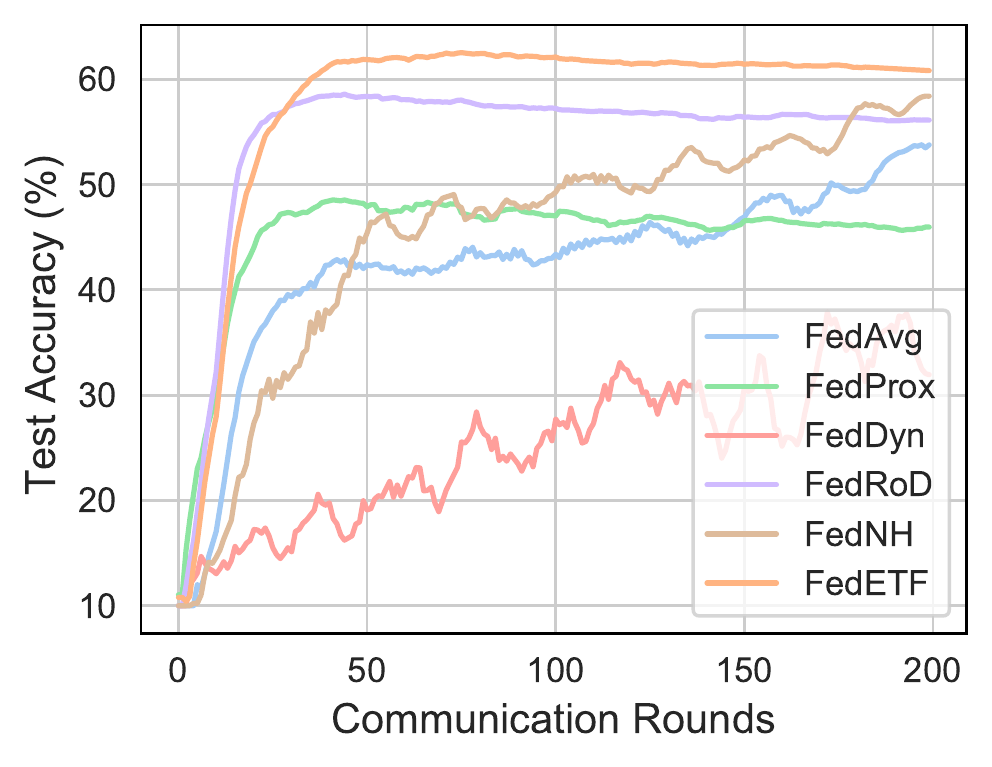}}
\subfigure[]{\includegraphics[width=0.5\columnwidth]{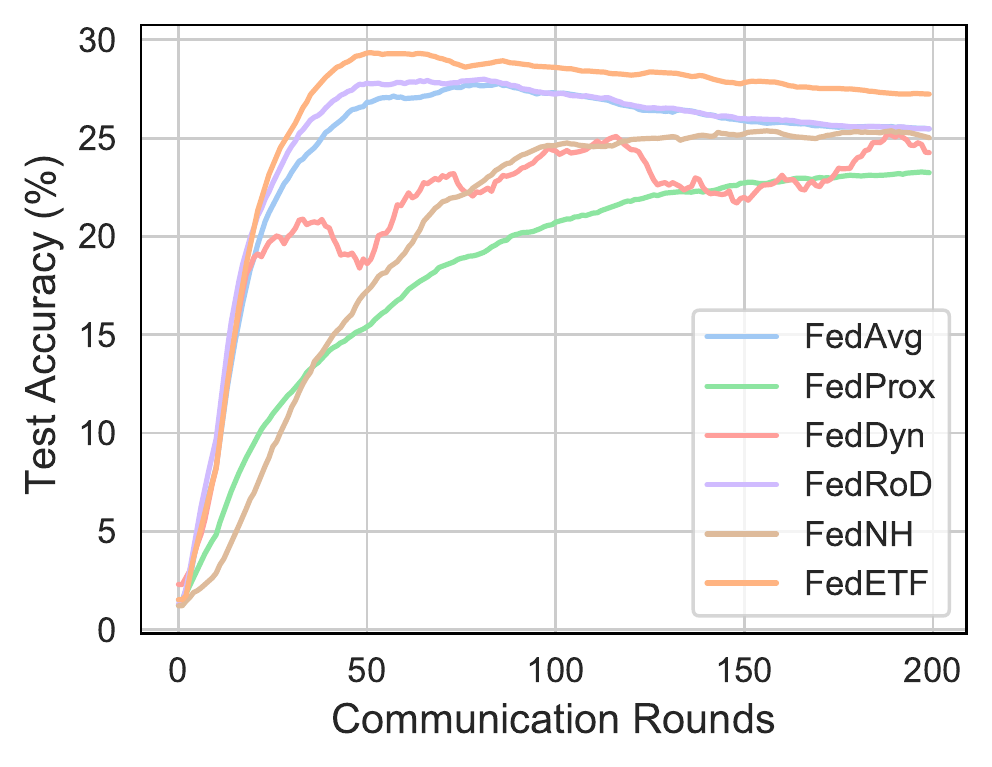}}
\caption{\textbf{Global models' test accuracy curves of the methods.} (a) CIFAR-10 with $\alpha=0.05$. (b) CIFAR-100 with $\alpha=0.05$. (c) CIFAR-10 with $\alpha=0.1$. (d) CIFAR-100 with $\alpha=0.1$. 
}
\label{fig:acc_curves}
\end{figure*}

\subsection{Personalized Adaptation by Local Finetuning}
\label{sect:fedetf_personalization}
As discovered in \cite{chen2021bridging}, local adaptation of a more generalized global model will reach stronger personalization. We will also verify this finding in our \FedETF. After the FL training, we obtain a global model $\bw^g$ with better generalization, and we use $\bw^g$ as the initialization in each client for personalization. We will show that by our tailored local finetuning of $\bw^g$, we will also reach the state-of-the-art in personalization. 

Our personalized local finetuning consists of two parts: \textit{local feature adaptation} and \textit{classifier finetuning}. 
In the \textit{local feature adaptation}, we fix the projection layer and ETF classifier and finetune the feature extractor to let the feature extractor be more customized to the features of clients' local data. 
In the \textit{classifier finetuning period}, we finetune the ETF classifier and projection layer alternately for several iterations to make the classifier more biased to the local class distributions. We note that the simplex ETF is not an ideal classifier for local personalization, since the clients may have imbalanced class distributions or even have missing classes. Biased classifiers are needed for personalization to take the local class distributions as prior knowledge and maximize the prediction likelihood. We alternately finetune the ETF classifier and projection layer to make the projected features and classifier vectors converge to be aligned. 

The process of local finetuning is illustrated in Figure \ref{fig:fedetf_personalization_framework}. The learned model parameters in the personalization stage are $\bw=\{\bu, \bp, \beta, \mathbf{V}_{ETF}\}$, and we split $\bw$ into the tuned parameters $\hat{\bw}$ and the fixed parameters $\overline{\bw}$. When finetune the feature extractor, $\hat{\bw}=\{\bu, \beta\},~ \overline{\bw}=\{\bp, \mathbf{V}_{ETF}\}$; when finetune the ETF classifier, $\hat{\bw}=\{\mathbf{V}_{ETF}, \beta\},~ \overline{\bw}=\{\bu, \bp\}$; when finetune the projection layer, $\hat{\bw}=\{\bp, \beta\},~ \overline{\bw}=\{\bu, \mathbf{V}_{ETF}\}$. We use the vanilla CE loss without balanced softmax in each stage of finetuning.
\begin{align}
    \ell^p(\hat{\bw},\overline{\bw}; \bx, y) = -\log\frac{\exp(\beta\cdot\bv_{y}^T\bmu)}{\sum_{c\in[C]}\exp(\beta\cdot\bv_{c}^T\bmu)}.
\end{align}
The learning objective in each stage is shown as follows.
\begin{align}
    \bw_k^p &= \{\overline{\bw}, \arg\min_{\hat{\bw}} \mathcal{L}_k^p(\hat{\bw})\}, \\
    \text{where }\mathcal{L}_k^p(\hat{\bw}) &= \frac{1}{n_k}\sum_{(\bx_i, y_i)\in \DD_k}\ell^p(\hat{\bw},\overline{\bw}; \bx_i, y_i).
\end{align}
Personalization will be reached after several iterative stages of finetuning in Figure \ref{fig:fedetf_personalization_framework}. The pseudo code of the personalized local finetuning is shown in Algorithm 2 in Appendix.

\section{Experiments and Results} \label{sect:exp}
\subsection{Settings}
\noindent\textbf{Datasets and Models.} Following previous works \cite{dai2022tackling,lin2020ensemble}, we use three vision datasets to conduct experiments: CIFAR-10 \cite{krizhevsky2009learning}, CIFAR-100 \cite{krizhevsky2009learning}, and Tiny-ImageNet \cite{deng2009imagenet,le2015tiny}. Tiny-ImageNet is a subset of ImageNet with 100k samples of 200 classes. Following \cite{li2018visualizing}, we adopt ResNet20 \cite{li2018visualizing,he2016deep} for CIFAR-10/100 and use ResNet-18 \cite{li2018visualizing,he2016deep} for Tiny-ImageNet. We use a linear layer as the classifier for the baselines and as the projection layer for our method.

\noindent\textbf{Compared Methods.} We take three lines of methods as baselines. 
\textit{\textbf{1) Classical FL with Non-IID data:}} \FedAvg \cite{mcmahan2017communication} with vanilla local training, a simple but strong baseline; \FedProx \cite{li2020federated}, FL with proximal regularization at clients; \FedDyn \cite{acar2020federated}, FL based on dynamic regularization. 
\textit{\textbf{2) Personalized FL:}} \Ditto \cite{li2021ditto}, personalization through separated local models; \FedRep \cite{collins2021exploiting}, personalization by only aggregating feature extractors; \FedRoD \cite{chen2021bridging}, personalization through decoupling models. 
\textit{\textbf{3) FL methods most relevant to ours:}} \CCVR \cite{luo2021no}, FL with classifier retraining; \FedProto \cite{tan2022fedproto}, FL with only prototype sharing; \FedRoD \cite{chen2021bridging}, generalization through decoupling and balanced softmax loss; \FedNH \cite{dai2022tackling}, FL with smoothing aggregation of prototypical classifiers. 

\begin{table*}[!t]
    \footnotesize
    \centering
    \caption{\textbf{Results (\%) under various numbers of clients with partial client sampling.} The dataset is CIFAR-10 with Non-IID $\alpha=0.1$.}
    \setlength\tabcolsep{10.76pt}
    \begin{tabular*}{0.99\linewidth}{l|cc|cc|cc|cc}
    \toprule
    Number of Clients&\multicolumn{4}{c}{50}&\multicolumn{4}{c}{100}\\
    \cmidrule(lr){1-5}
    \cmidrule(lr){6-9}
    Sampling Rate &\multicolumn{2}{c}{0.4}&\multicolumn{2}{c}{0.6}&\multicolumn{2}{c}{0.4}&\multicolumn{2}{c}{0.6}\\
    \midrule
    Methods/Metrics &General.&Personal.&General.&Personal.&General.&Personal.&General.&Personal.\\
    \midrule
    \textsc{FedAvg} \cite{mcmahan2017communication}&38.13{\tiny±5.12}	&77.28{\tiny±2.17}	
    &42.68{\tiny±6.28}	&74.99{\tiny±2.34}
    &42.15{\tiny±1.61}	&71.52{\tiny±1.88}	&41.42{\tiny±3.31}	&70.40{\tiny±2.13}\\
    \midrule
    \textsc{CCVR} \cite{luo2021no}&44.59{\tiny±11.4}	&\textbf{78.93{\tiny±3.26}}	&52.49{\tiny±6.73}	&\textbf{82.33{\tiny±1.72}}	&50.07{\tiny±0.80}	&\textbf{76.27{\tiny±2.08}}	&50.41{\tiny±3.93}	&\textbf{77.27{\tiny±1.22}}\\
    \textsc{FedRoD} \cite{chen2021bridging}&\textbf{55.84{\tiny±3.96}}	&76.60{\tiny±0.13}	
    &\textbf{53.04{\tiny±2.54}}	&74.42{\tiny±1.99}
    &\textbf{52.62{\tiny±1.68}} &71.27{\tiny±0.69}	&\textbf{52.34{\tiny±0.11}}	&72.41{\tiny±0.74}\\
    \textsc{FedNH} \cite{dai2022tackling}&39.97{\tiny±6.90}	&76.59{\tiny±0.59}	
    &45.36{\tiny±3.58}	&78.17{\tiny±1.15}
    &42.77{\tiny±0.65} &73.47{\tiny±1.38}	&45.85{\tiny±2.98}	&73.15{\tiny±0.95}\\
    \midrule
    \rowcolor{gray!20}\textbf{Our \textsc{FedETF}}&\textbf{58.05{\tiny±4.63}}	&\textbf{85.82{\tiny±0.86}}	&\textbf{58.75{\tiny±1.72}}	&\textbf{85.05{\tiny±0.87}}
    &\textbf{56.67{\tiny±0.88}}	&\textbf{83.47{\tiny±0.45}}	&\textbf{55.96{\tiny±0.23}}	
    &\textbf{83.38{\tiny±0.72}}\\
    \bottomrule
    \end{tabular*}
    \label{table:partial_select}
    \vspace{-0.2cm}
\end{table*}

\begin{table*}[!t]
    \footnotesize
    \centering
    \caption{\textbf{Results (\%) under different local epochs ($E$).} The dataset is CIFAR-10 with Non-IID $\alpha=0.1$.}
    \setlength\tabcolsep{10.76pt}
    \begin{tabular*}{0.99\linewidth}{l|cc|cc|cc|cc}
    \toprule
    $E$ &\multicolumn{2}{c}{1}&\multicolumn{2}{c}{2}&\multicolumn{2}{c}{4}&\multicolumn{2}{c}{8}\\
    \midrule
    Methods/Metrics &General.&Personal.&General.&Personal.&General.&Personal.&General.&Personal.\\
    \midrule
    \textsc{FedAvg} \cite{mcmahan2017communication}&45.75{\tiny±1.97}	&74.18{\tiny±2.22}	
    &43.02{\tiny±2.16}	&77.45{\tiny±0.79}
    &36.30{\tiny±2.88}	&80.66{\tiny±1.37}	&32.58{\tiny±4.87}	&84.24{\tiny±0.92}\\
    \midrule
    \textsc{CCVR} \cite{luo2021no}&59.82{\tiny±4.35}	&\textbf{79.83{\tiny±1.66}}	&53.73{\tiny±8.48}	&80.43{\tiny±2.54}	&55.73{\tiny±4.00}	&81.83{\tiny±1.32}	&\textbf{55.00{\tiny±2.29}}	&\textbf{85.61{\tiny±1.12}}\\
    \textsc{FedRoD} \cite{chen2021bridging}&\textbf{60.34{\tiny±3.22}}	&77.10{\tiny±1.97}	
    &\textbf{56.74{\tiny±5.59}}	&76.96{\tiny±4.26}
    &\textbf{57.85{\tiny±5.22}} &81.08{\tiny±1.74}	&50.63{\tiny±9.49}	&84.75{\tiny±0.96}\\
    \textsc{FedNH} \cite{dai2022tackling}&39.14{\tiny±9.56}	&77.27{\tiny±1.53}	
    &45.13{\tiny±1.60}	&\textbf{81.84{\tiny±0.88}}
    &40.28{\tiny±2.78} &\textbf{82.90{\tiny±1.00}}	&39.18{\tiny±3.66}	&83.95{\tiny±1.23}\\
    \midrule
    \rowcolor{gray!20}\textbf{Our \textsc{FedETF}}&\textbf{62.76{\tiny±2.90}}	&\textbf{88.00{\tiny±0.65}}	&\textbf{62.34{\tiny±4.10}}	&\textbf{88.20{\tiny±0.93}}
    &\textbf{61.78{\tiny±3.21}}	&\textbf{88.46{\tiny±0.75}}	&\textbf{54.23{\tiny±10.8}}	
    &\textbf{88.40{\tiny±0.98}}\\
    \bottomrule
    \end{tabular*}
    \label{table:local_epoch}
\end{table*}

\noindent\textbf{Client Settings.} We adopt the Dirichlet sampling to generate Non-IID data for each client. We note that the Dirichlet-sampling-based data heterogeneity is widely used in FL literature \cite{lin2020ensemble,chen2021bridging,dai2022tackling,luo2021no}. It considers a class-imbalanced data heterogeneity, controlled by hyperparameter $\alpha$, and smaller $\alpha$ refers to more Non-IID data of clients. 
In our experiments, we evaluate the methods under strong Non-IID settings with $\alpha \in \{0.1,0.05\}$ \cite{luo2021no,zhang2022federated}. 
If without mentioning otherwise, the number of clients $K=20$ and we adopt full client participation. 
For CIFAR-10 and CIFAR-100 the number of local epochs $E=3$ and the number of communication rounds $T=200$, while for Tiny-ImageNet, considering the high computation costs, we set $E=1$ and $T=50$. 

\noindent\textbf{Evaluation Metrics.} We test the generalization of the aggregated global model (General.) and the personalization of clients' local models (Personal.). Generalization performance is validated on the balanced testset of each dataset after the global model is generated on the server. For each client, we split 70\% of the local data for the trainset and 30\% for the testset. Following \cite{chen2021bridging}, we validate the personalization performance on each client's local testset after the local training and average the personalized accuracies. 

\noindent\textbf{Implementation.} In all the experiments, we conduct three trials for each setting and present the mean accuracy and the standard deviation in the tables. For more implementation details, please refer to the Appendix. 

\subsection{Main Results}
\noindent\textbf{Results under various vision datasets and data heterogeneity.} Table \ref{table:first_table} shows the results of all methods on three vision datasets with Non-IID $\alpha \in \{0.1, 0.05\}$. \textit{Our method achieves state-of-the-art performances in 11 out of 12 settings in both generalization and personalization. It is notable that except for our \FedETF, there is no comparable baseline that can achieve high results in all datasets.} Generally, our method has more significant improvement under more heterogeneous settings ($\alpha=0.05$), especially in CIFAR-10. We also visualize the learning curves in Figure \ref{fig:acc_curves}. \textit{Our \FedETF not only has higher accuracies but also has faster and more steady convergence. }

For generalization, in most cases, \FedRoD can improve the accuracies upon \FedAvg, which showcases the effectiveness of the balanced loss. However, the balanced loss cannot thoroughly solve the classifier bias problem, and \textit{our method \FedETF which adopts the optimal classifier structure has large-margin gains over \FedRoD}. We notice that the classifier retraining algorithm \CCVR is not effective in all cases, which indicates that the retraining method is not practical enough for solving classifier biases. 

For personalization, we find the personalized FL \FedRep and \FedRoD are strong baselines. Compared with these personalized FL methods, our \FedETF also reaches the state-of-the-art in almost all cases. \textit{The success of \FedETF in personalization is the result of training a better generalized global model and the effective local adaption of such a global model.} 

\begin{figure*}[!h]
\centering
\subfigure[]{\includegraphics[width=0.52\columnwidth]{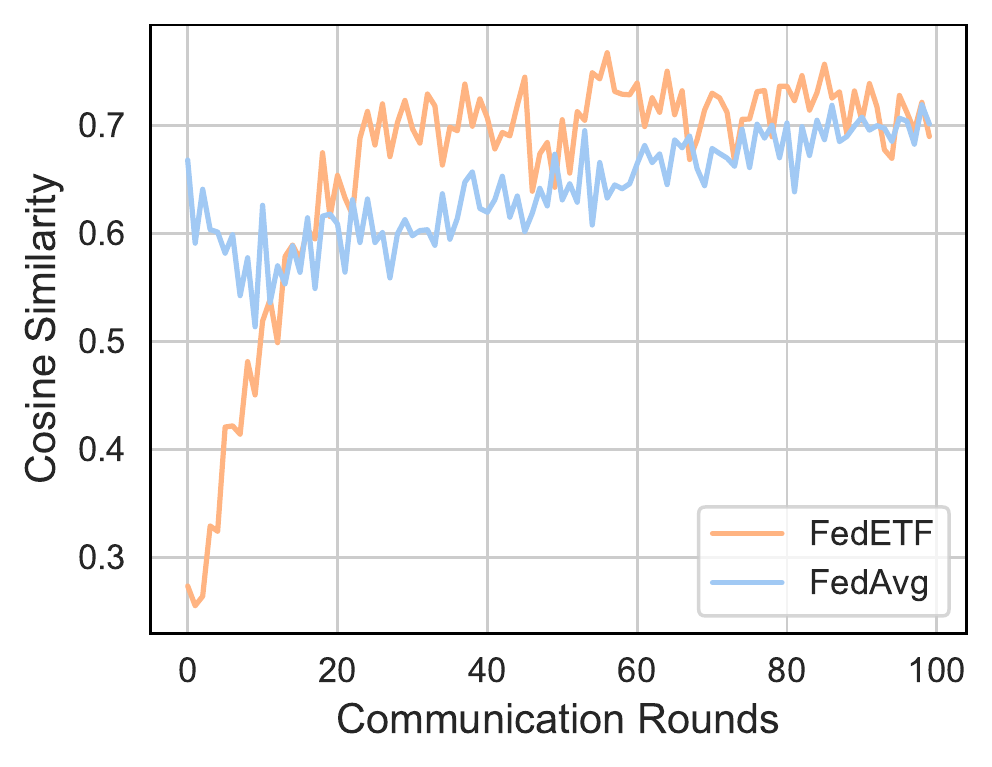}}
\subfigure[]{\includegraphics[width=0.52\columnwidth]{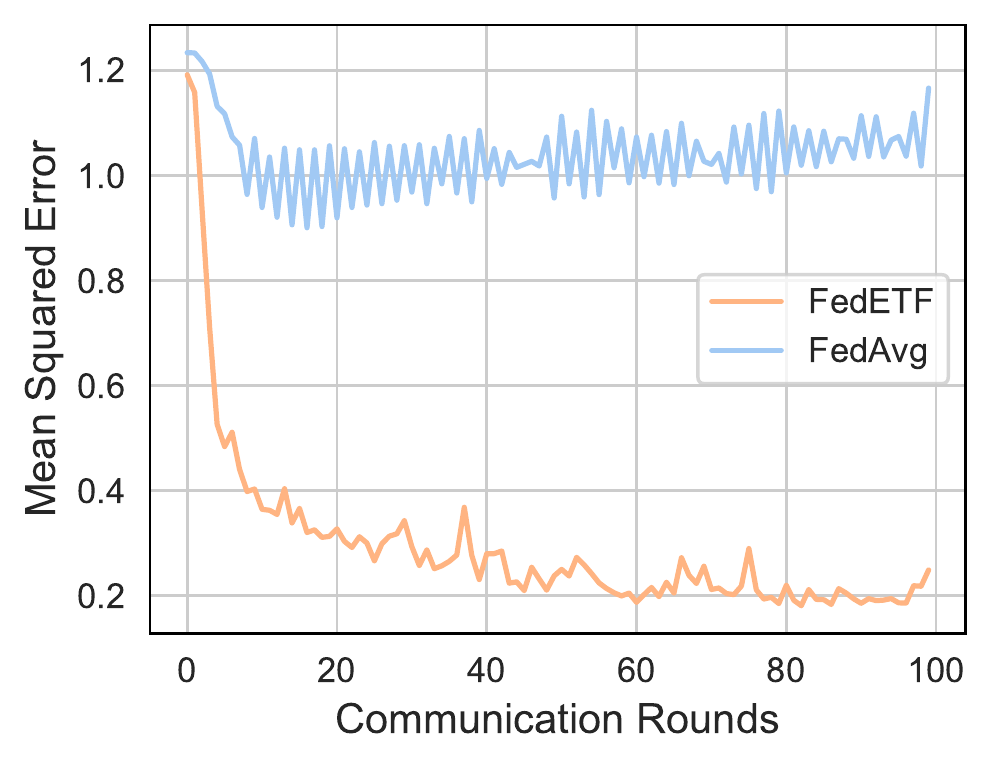}}
\subfigure[]{\includegraphics[width=0.52\columnwidth]{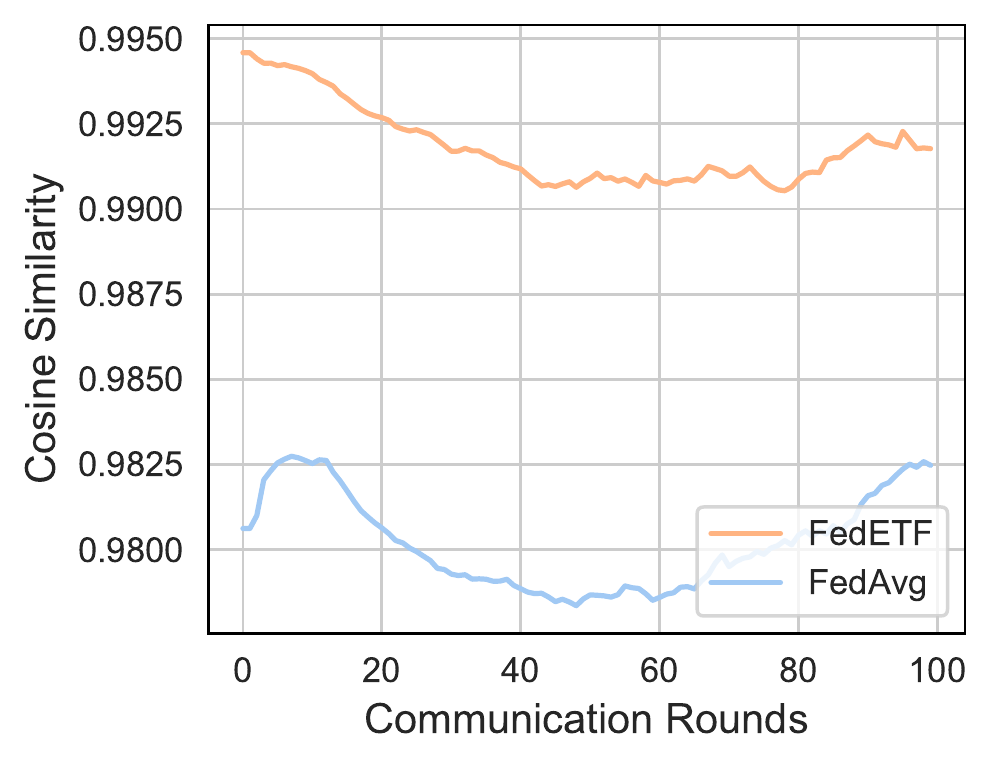}}
\caption{\textbf{Understanding feature alignment and neural collapse of \textsc{FedETF}.} Test accuracy of final global models: \FedAvg 29.73\%, \FedETF 54.95\%. Experiments on CIFAR-10 with $\alpha=0.05$. (a) Feature prototype consistency of clients' local models, higher values mean better feature alignment. (b) Neural collapse error of the aggregated global model, lower values mean greater neural collapse. (c) Local model consistency: averaged pair-wise cosine similarities of clients' local models, higher values mean smaller model drifts.}
\label{fig:understanding_fedetf_nc}
\end{figure*}

\noindent\textbf{Results under different $K$ with partial client sampling.} 
We select the best baselines in Table \ref{table:first_table} and conduct experiments on CIFAR-10 under various numbers of clients $K$ with partial client sampling in Table \ref{table:partial_select}. We set $K \in \{50,100\}$ and set the sampling rate as 0.4 and 0.6. To ensure fair comparisons, we randomly generate and save a sampling list in advance and let every method load the same sampling list during training. It is obvious that our \FedETF is the best method in both generalization and personalization. Compared with \FedAvg, the advantage of our method is more dominant under the smaller sampling rate, i.e. 0.4. With respect to generalization, when $K=50$, the improvement is 5.2\% for the sampling rate 0.4, and 3.8\% for the sampling rate 0.6. \textit{It indicates that the fixed ETF classifier is more robust than the learnable classifier to tackle \textbf{system heterogeneity}.} Moreover, our method has smaller variances of accuracies, which also verifies its robustness.

\noindent\textbf{Results under different local epochs $E$.} In Table \ref{table:local_epoch}, generally, when the number of local epochs varies, our \FedETF also achieves constantly state-of-the-art performances. 
For \FedAvg and \FedRoD, when $E$ is larger, the generalization performance will decline. \textit{However, for our \FedETF, the declines are weaker, showing its robustness and effectiveness. For personalization, \FedETF has more steady and promising results under different $E$.}

\begin{table}[t]\footnotesize
\centering
\caption{\textbf{Evaluation (\%) of \textsc{FedETF} using different model architectures.} The dataset is CIFAR-10 with Non-IID $\alpha=0.1$.}
\begin{tabular*}{0.99\linewidth}{lcc|cc}
    \toprule
    Methods& \multicolumn{2}{c}{\textsc{FedAvg}} & \multicolumn{2}{c}{\textsc{FedETF}} \\
    \midrule
    Models/Metrics& General.& Personal.& General.& Personal.\\
    \midrule
     DenseNet121   & 62.87{\tiny±2.23}  & 85.66{\tiny±3.70}& \textbf{74.92{\tiny±2.75}}  & \textbf{91.83{\tiny±0.67}} \\
     MobileNetV2   & 43.20{\tiny±4.45}  & 87.07{\tiny±1.04}& \textbf{57.43{\tiny±12.0}}  & \textbf{89.88{\tiny±0.40}} \\
     EfficientNet   & 35.92{\tiny±4.47}  & 84.69{\tiny±1.26}& \textbf{56.70{\tiny±5.52}}  & \textbf{87.50{\tiny±0.74}} \\
     \midrule
     ResNet20   & 52.76{\tiny±6.08}  & 83.85{\tiny±0.89}& \textbf{59.56{\tiny±1.84}}  & \textbf{87.89{\tiny±1.19}} \\
     ResNet32   & 53.22{\tiny±7.73}  & 82.90{\tiny±4.31}& \textbf{60.71{\tiny±2.67}}  & \textbf{87.97{\tiny±1.17}} \\
     ResNet56   & 57.09{\tiny±6.10}  & 83.70{\tiny±4.65}& \textbf{60.44{\tiny±3.57}}  & \textbf{88.23{\tiny±0.99}} \\
     WRN56\_4   & 63.64{\tiny±4.91}  & 86.76{\tiny±1.08}& \textbf{66.30{\tiny±3.88}}  & \textbf{89.94{\tiny±0.52}} \\
    \bottomrule
  \end{tabular*}
\label{table:model_architecture}  
\end{table}

\subsection{Understanding \textbf{\FedETF}}
\noindent\textbf{Evaluation using different model architectures.} 
We consider two scenarios: \textbf{\textit{1) Different backbones.}} DenseNet121 \cite{huang2017densely}, MobileNetV2 \cite{howard2018inverted,sandler2018mobilenetv2}, and EfficientNet \cite{tan2019efficientnet}. \textbf{\textit{2) Deeper and wider models.}} 
Deeper models: ResNet20, ResNet32, and ResNet56 \cite{li2018visualizing}  (the larger number refers to the deeper model); wider models: ResNet56 and WRN56\_4 \cite{li2018visualizing} (WRN: the abbreviation for Wide ResNet). The results are shown in Table \ref{table:model_architecture}. \textit{Our method can also improve performance under various model architectures.} For models with different depths, we observe that \FedETF has larger superiority in shallower models. 
Specifically, \FedETF can release the full potential of ResNet20 (shallower and smaller model) to let it has even better performance than \FedAvg with ResNet56 (deeper and larger model). \textit{It showcases the applicability of \FedETF that it can enable smaller models to have better performances than the larger ones, saving both computation and communication costs in FL.}

\noindent\textbf{Why does \FedETF work well?} We explore how the features are learned in \FedETF compared with \FedAvg. We first examine the feature alignment of local models. In each round, after local training, we compute class prototypes (feature mean of each class) in each client and calculate the cosine similarities of clients' class-wise prototypes, which is analogous to NC1 in Section \ref{sect:pre_nc}. Then we average all the cosine similarities to indicate the feature alignment, a larger value reveals more aligned clients' features. The results are in Figure \ref{fig:understanding_fedetf_nc} (a). \textit{It shows that only after a few rounds, \FedETF has constantly stronger clients' feature alignment than \FedAvg, showing the fixed ETF classifier is effective to align local features.}

We also study whether \FedETF can help the global model reach neural collapse in terms of NC2 in Figure \ref{fig:understanding_fedetf_nc} (b). 
In each round, we first compute the class prototypes of the global model and calculate the pair-wise cosine similarities of these prototypes. In neural collapse optimality (Definition \ref{def:ETF}), the pair-wise cosine similarities of prototypes are $-\frac{1}{K-1}$. Hence, we calculate the mean square error between the global model's cosines and $-\frac{1}{K-1}$ to indicate the neural collapse error. Results display that \FedETF has a much smaller neural collapse error than \FedAvg, and the error of \FedETF is decreasing along the training. \textit{It indicates that \FedETF can help the global model reach neural collapse.} Note that \FedAvg has 29.73\% in global test accuracy while \FedETF has 54.95\%. 
\textit{It also verifies that the global model's generalization is connected with neural collapse optimality in FL, which is consistent with the observations in centralized training \cite{li2022principled}.} 

Additionally, we study why \textsc{FedETF} works well from the perspective of model drift (local model consistency) in Figure \ref{fig:understanding_fedetf_nc} (c). Results demonstrate that \textsc{FedETF} has stably higher local model consistency than \textsc{FedAvg} throughout the training, indicating lower model drift. It is consistent with the intuition that the fixed ETF classifier can reduce the model drifts of feature extractors, resulting in better aggregation and convergence.

\begin{figure}[!t]
\centering
\subfigure[]{\includegraphics[width=0.48\columnwidth]{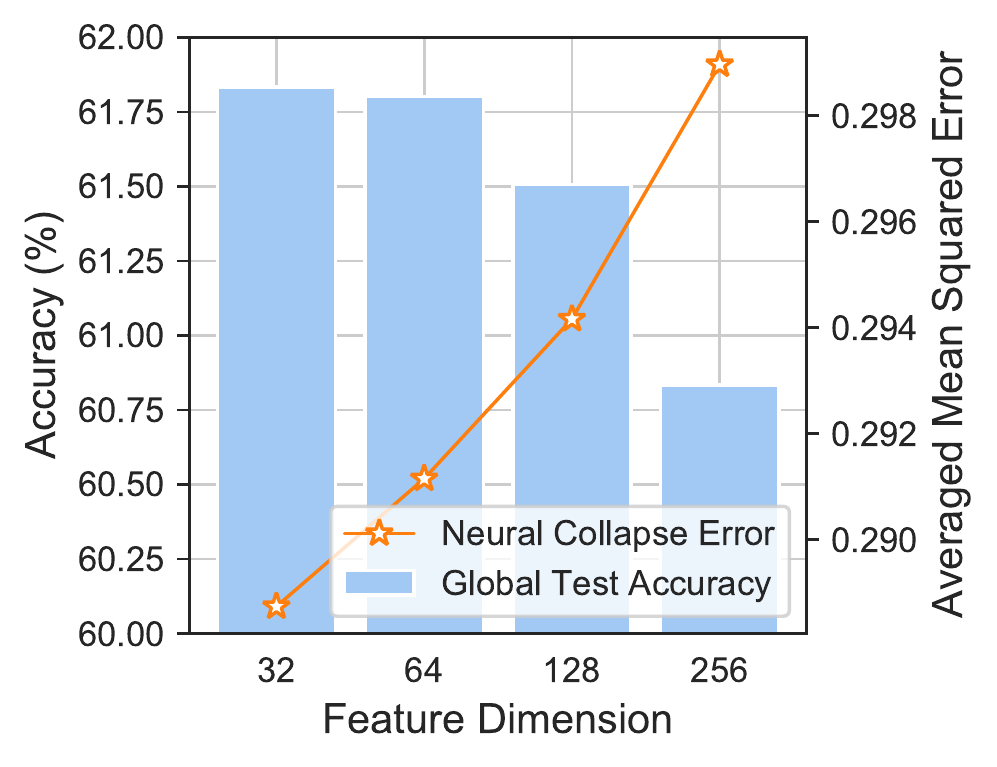}}
\subfigure[]{\includegraphics[width=0.48\columnwidth]{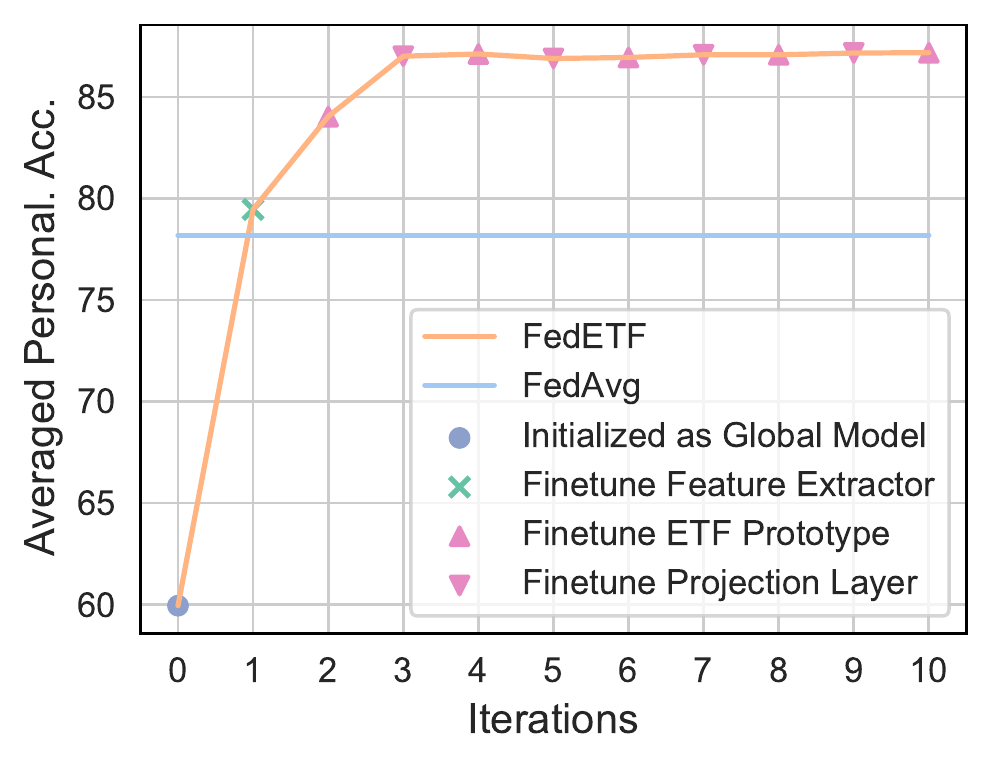}}
\caption{\textbf{Understanding feature dimension and local personalization in \FedETF.} Experiments on CIFAR-10 with $\alpha=0.1$. (a) How feature dimension affects \FedETF's generalization and neural collapse. (b) How personalization is reached in each iteration of \FedETF's local finetuning. }
\label{fig:understanding_fedetf_featuredim_personalization}
\end{figure}

\noindent\textbf{How feature dimension affects \FedETF.} We analyse how the feature dimension $d$ of ETF affects \FedETF's performance on CIFAR-10 in Figure \ref{fig:understanding_fedetf_featuredim_personalization} (a). We find that smaller $d$ will cause smaller neural collapse errors and are slightly beneficial to generalization. Random high-dimensional vectors are more prone to be orthogonal, so we suppose that prototypes in higher dimensions are more likely to be orthogonal. In CIFAR-10, the number of classes $K=10$, and the ETF angles are obtuse with -0.11 pair-wise cosines. Therefore, it is hard for high-dimensional features to collapse into the obtuse angle's structure. \textit{We suggest setting the feature dimension $d$ in \FedETF according to the number of classes $C$.} If $C$ is small, it also requires a relatively small $d$ to improve neural collapse.

\noindent\textbf{How personalization is reached during local finetuning.} We visualize the averaged personalized accuracies of different iterations during \FedETF's local finetuning in Figure \ref{fig:understanding_fedetf_featuredim_personalization}~(b). At first, when clients' local models are initialized as the final global model, the personalization is poor. After finetuning the feature extractor, \FedETF has better results than the baseline \FedAvg with finetuning. Then alternatively finetuning the ETF classifier and projection layer further improves the personalization and makes the accuracy converge to a higher point. 

\subsection{Ablation Study}
We conduct the ablation study of \FedETF in terms of generalization\footnote{Figure \ref{fig:understanding_fedetf_featuredim_personalization} (b) can be viewed as the ablation study of the personalized local finetuning stage.} on CIFAR-10 with Non-IID $\alpha \in \{0.1, 0.05\}$. \textit{It is found that every module in \FedETF plays an important role and the modules strengthen each other to realize better performances.} If taking one module off, the performance will meet severe declines, but the results are still better than \FedAvg in general. We notice the balanced loss module is more important under a more heterogeneous environment, and this observation is consistent with previous works in neural collapse \cite{yang2022we} and FL \cite{chen2021bridging}. It is also notable to emphasize the significance and necessity of the projection layer. Our method without a projection layer only has marginal gains over \FedAvg. We also find that \FedNH has relatively poor performances in Table \ref{table:first_table}, especially on CIFAR-100 and Tiny-ImageNet, and we suppose the main cause may be that \FedNH does not adopt a projection layer to map the raw features into a space where neural collapse is more prone to happen. Moreover, the learnable temperature is also crucial for \FedETF to adaptively adjust the softmax temperature so as to meet the learning dynamics of feature representations.

\section{Conclusion}
In this paper, we fundamentally solve the classifier biases caused by data heterogeneity in FL by proposing a neural-collapse-inspired solution. Specifically, we employ a simplex ETF as a fixed classifier for all clients during federated training, which allows them to learn unified and optimal feature representations. Afterward, we introduce a novel finetuning strategy to enable clients to have more personalized local models. 
Our method achieves the state-of-the-art performance regarding both generalization and personalization compared to strong baselines, as shown by extensive experimental results on CIFAR-10/100 and Tiny-ImageNet. Furthermore, we gained insights into understanding the effectiveness and applicability of our approach.

\begin{table}[!t]\footnotesize
\centering
\caption{\textbf{Ablation study of \textsc{FedETF} in terms of global model's generalization.} The dataset is CIFAR-10.}
\setlength\tabcolsep{11.55pt}
\begin{tabular*}{0.99\linewidth}{lcc}
    \toprule
    Methods/NonIID($\alpha$) & 0.1 & 0.05\\
    \midrule
     \textsc{FedAvg}   & 43.75{\tiny±0.42}  & 38.47{\tiny±1.95} \\
     \midrule
    Ours w/o Projection Layer &44.92{\tiny±6.22} &  41.91{\tiny±1.47}\\
    Ours w/o Balanced Loss &46.06{\tiny±0.75} &  37.63{\tiny±4.45} \\
    Ours w/o Learnable Temperature  &49.80{\tiny±4.52} & 46.07{\tiny±1.61} \\
    \midrule
     \rowcolor{gray!20}Ours    & \textbf{56.46{\tiny±4.18}} & \textbf{53.98{\tiny±1.29}}\\
    \bottomrule
  \end{tabular*}
\label{table:ablation}  
\end{table}

\subsection*{Acknowledgments} This work was supported by the National Key Research and Development Project of China (2021YFC3340300), National Natural Science Foundation of China (U19B2042), the Zhejiang Provincial Key Research and Development Project (2022C03106), University Synergy Innovation Program of Anhui Province (GXXT-2021-004), Academy Of Social Governance Zhejiang University, Fundamental Research Funds for the Central Universities (226-2022-00064). This work was supported in part by the National Key R\&D Program of China (Project No.\ 2022ZD0115100), the Research Center for Industries of the Future (RCIF) at Westlake University, and Westlake Education Foundation.


\newpage
\appendix
{\LARGE \textbf{Appendix}}

\section{Pseudo Codes of the Proposed Method}
To better present our approach and demonstrate the workflow, we give the pseudo codes of \FedETF. The pseudo code of the federated learning (FL) training is shown in Algorithm \ref{algorithm1}, and the pseudo code of the personalized local finetuning is shown in Algorithm \ref{algorithm2}.

\begin{algorithm}[th]\label{algorithm1}
\caption{ \FedETF FL Training }
\KwIn{Clients $\{1,\dots,K\}$, communication round $T$, local epoch $E$, initial model $\bw^1=\{\bu, \bp, \beta\}$, feature dimension $d$, balanced loss hyperparameter $\gamma$.}
\KwOut{Gloabl model $\bw^g$.}
Synthesize a simplex ETF $\mathbf{V}_{ETF} \in \mathbb{R}^{d\times C}$ by Eq. (3) as the fixed classifier for all clients;\\
\For{$t=1,\dots, T$}{
\For{{\rm client} $k =1,\dots, K$ \textbf{in parallel}}{
$\bw_k^{t} \leftarrow \bw^{t}$; \\
\For{{\rm local epoch} $e=1,\dots, E$}{
Obtain $\mathcal{L}_k^g$ by Eq. (6,~7,~8);
$\bw_k^{t} \leftarrow \bw_k^{t} -\eta \nabla \mathcal{L}_k^g(\bw_k^{t})$;
}
}
The server updates $\bw^{t+1}$ by Eq. (2).
}
The final global model $\bw^g$ = $\bw^T$.
\end{algorithm}

\begin{algorithm}[th]
\caption{ \FedETF Personalized Finetuning }\label{algorithm2}
\KwIn{Clients $\{1,\dots,K\}$, iteration round $T_p$, epoch for each stage $E$, final global model $\bw^g=\{\bu, \bp, \beta, \mathbf{V}_{ETF}\}$.}
\KwOut{Personalized local models $\{\bw_k^p\}_{k=1}^{K}$.}
Assign the final global model $\bw^g$ as clients' initial local models.\\
\textit{Finetune the feature extractor.} \\
\For{{\rm client} $k =1,\dots, K$ \textbf{in parallel}}{
$\hat{\bw}_k=\{\bu, \beta\},~ \overline{\bw}_k=\{\bp, \mathbf{V}_{ETF}\}$; \\
\For{{\rm local epoch} $e=1,\dots, E$}{
Obtain $\mathcal{L}_k^p$ by Eq. (9,~10,~11);
$\hat{\bw}_k \leftarrow \hat{\bw}_k -\eta \nabla \mathcal{L}_k^p(\hat{\bw}_k)$;
}}
\For{$t=1,\dots, T_p$}{
\For{{\rm client} $k =1,\dots, K$ \textbf{in parallel}}{
\textit{Finetune the ETF classifier.} \\
$\hat{\bw}_k=\{\mathbf{V}_{ETF}, \beta\},~ \overline{\bw}_k=\{\bp, \bu\}$; \\
\For{{\rm local epoch} $e=1,\dots, E$}{
Obtain $\mathcal{L}_k^p$ by Eq. (9,~10,~11);
$\hat{\bw}_k \leftarrow \hat{\bw}_k -\eta \nabla \mathcal{L}_k^p(\hat{\bw}_k)$;
}
\textit{Finetune the projection layer.} \\
$\hat{\bw}_k=\{\bp, \beta\},~ \overline{\bw}_k=\{\mathbf{V}_{ETF}, \bu\}$; \\
\For{{\rm local epoch} $e=1,\dots, E$}{
Obtain $\mathcal{L}_k^p$ by Eq. (9,~10,~11);
$\hat{\bw}_k \leftarrow \hat{\bw}_k -\eta \nabla \mathcal{L}_k^p(\hat{\bw}_k)$;
}
}
}
The personalized local models are $\{\bw_k^p = \hat{\bw}_k \cup \overline{\bw}_k\}_{k=1}^{K}$.
\end{algorithm}

\begin{figure*}[t]
 \vspace{-1.1cm}
\label{fig:data_distr}
\centering
\includegraphics[width=0.9\columnwidth]{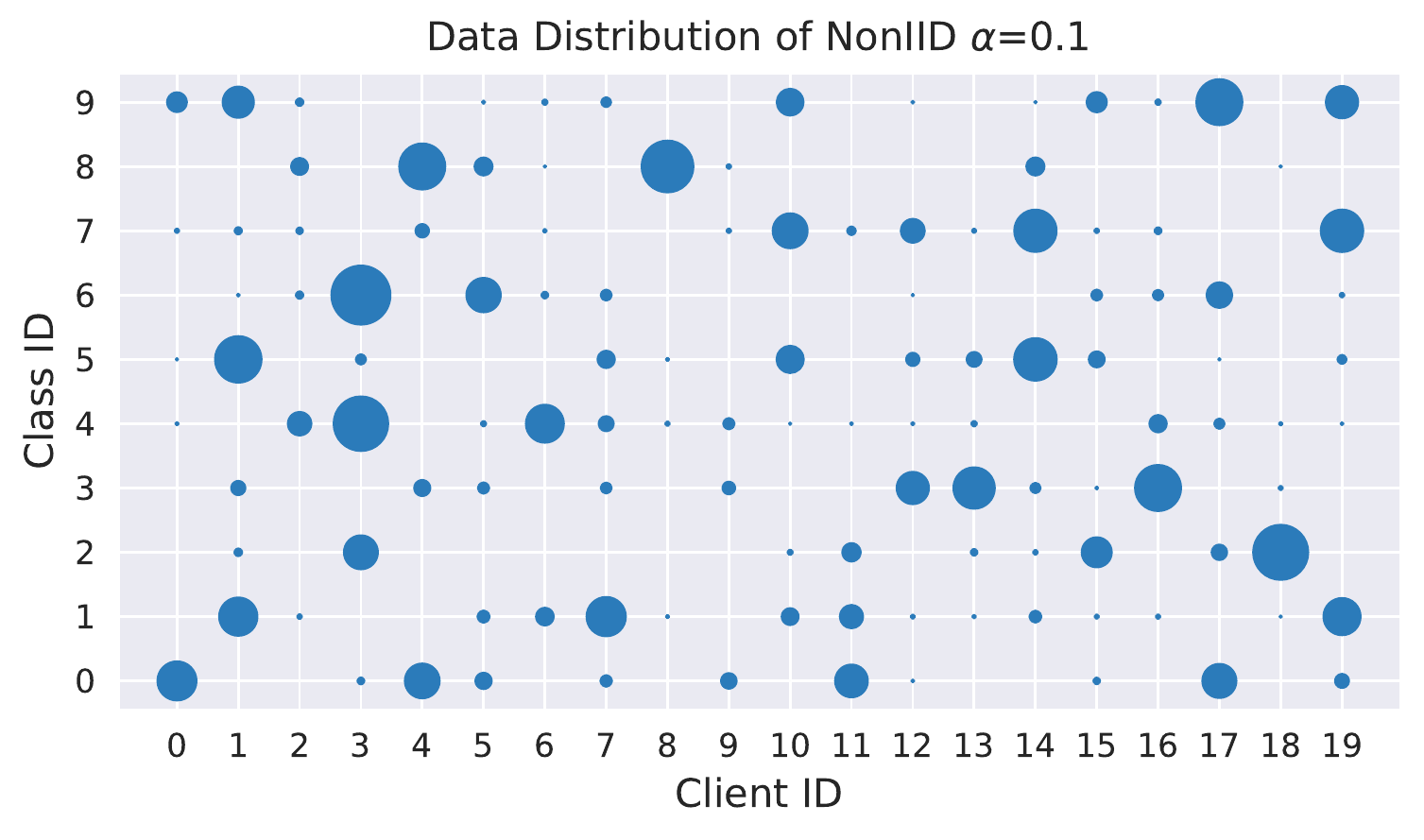}
\includegraphics[width=0.9\columnwidth]{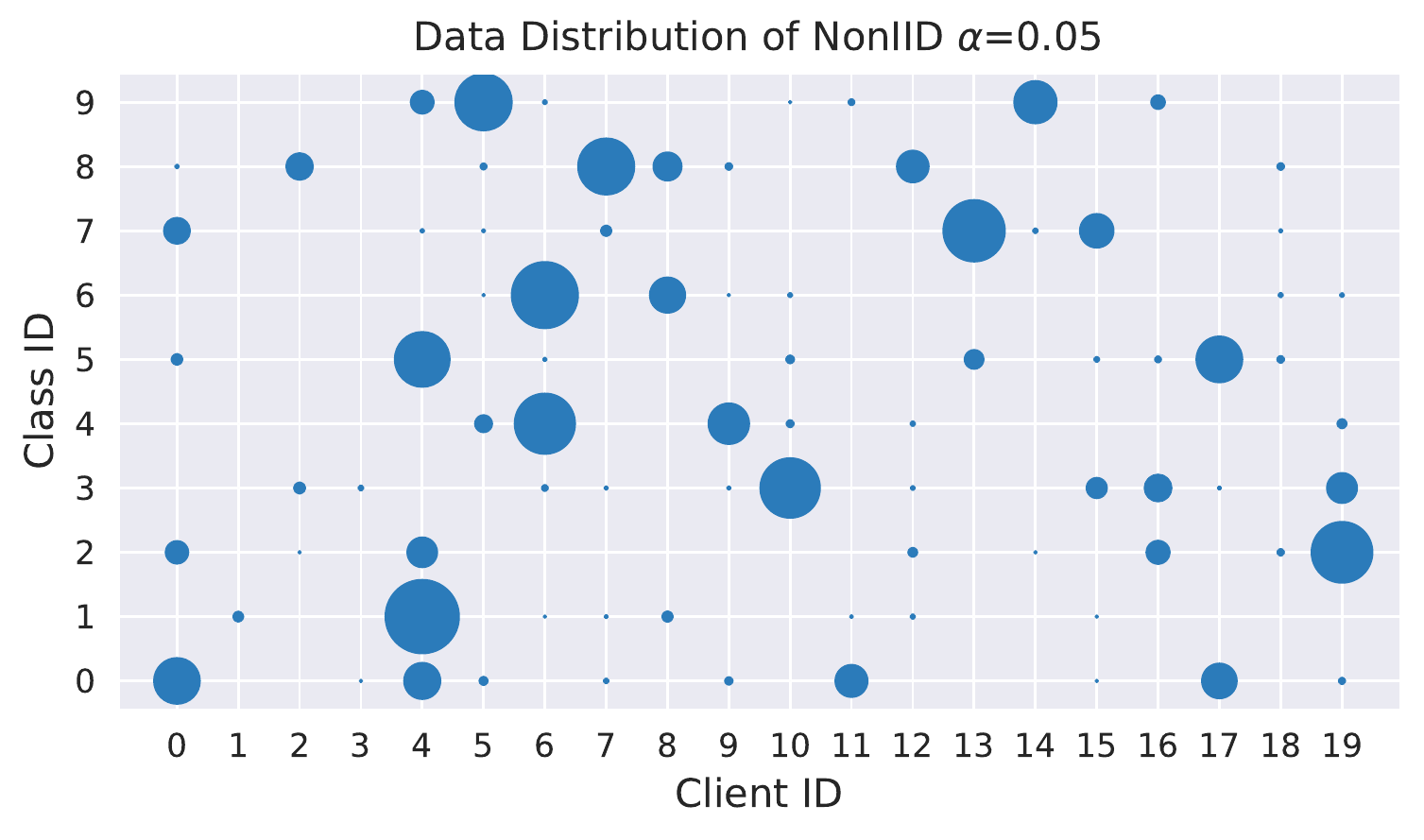}
\caption{ \textbf{Visualization of clients' data distributions.} Random seed is 8. Left: data distributions of Non-IID $\alpha=0.1$ with 20 clients. Right: data distributions of Non-IID $\alpha=0.05$ with 20 clients.}
\end{figure*}

\section{Implementation Details}
\noindent\textbf{Models and Data.} Our model implementations of the ResNet series and the DenseNet are referred from the codes of \cite{li2018visualizing}. The model implementation of the EfficientNet is referred from the official code in \cite{tan2019efficientnet}, and the implementation of the MobileNetv2 is referred from \cite{sandler2018mobilenetv2,howard2018inverted}. For the data, we use the Dirichlet-sampling-based data partition adopted in \cite{lin2020ensemble,chen2021bridging,dai2022tackling,luo2021no}. It considers a class-imbalanced data heterogeneity, controlled by hyperparameter $\alpha$, and smaller $\alpha$ refers to more Non-IID data of clients. When $\alpha < 1$, the data are considered to be rather Non-IID, which means that most of the training samples of one class are likely assigned to a small portion of clients \cite{chen2021bridging}. In our Dirichlet implementation, when $\alpha$ goes smaller, the number of samples in each client along with the class distribution of each client both become more heterogeneous, which is realistic in practical scenarios. 
We use the same Tiny-ImageNet dataset as in \cite{dai2022tackling}. 

\noindent\textbf{Local learning rate and optimizer.} For CIFAR-10 the local learning rate (LR) $\eta=0.04$, and for CIFAR-10 and Tiny-ImageNet, $\eta=0.01$. For clients, we use SGD optimizer with momentum 0.9 and weight decay $5\times10^{-4}$. Following \cite{DBLP:conf/nips/GhoshCYR20}, we adopt a learning rate decaying scheduler, which decays the local LR by 0.99 in each round. 

\noindent\textbf{Hyperparameters.} 
For \FedETF, we set the feature dimension to the number of classes, i.e. $d = C$; the initial temperature $\beta=1$; $\gamma=1$. 
We set $\mu_{FedProx} = 0.001$ in \FedProx and $\alpha_{FedDyn} = 0.01$ in \FedDyn as suggested in their official implementations or papers. For \Ditto, the learning setting of the personalized model is the same as the one of the global model. For \FedRep, the epoch number of the classifier training and the epoch number of the feature extractor training are the same and are set as $E$. For \FedRoD, we set $\gamma=1$. For \CCVR, the number of virtual features is 10 per class, and the number of classifier calibration training epochs is 100. For \FedNH, the smoothing hyperparameter $\rho=0.9$ as suggested in the paper \cite{dai2022tackling}. 

\noindent\textbf{Randomness.} We set the same random seeds for all methods in the same setting. The random seed list is $\{7, 8, 9, 10\}$. For the extremely Non-IID settings when $\alpha=0.05$, we use the random seeds that can ensure all clients can be assigned a proportion of training data (on the contrary, some random seeds will generate a data partition where particular clients have zero data samples). 

\noindent\textbf{Environments.} All experiments are conducted in PyTorch with Quadro RTX 8000 GPUs.

\section{Visualization}
\subsection{Visualization of clients' data distributions.} Here, we additionally visualize the clients' data distributions mainly adopted in the main paper. In the main paper, we adopt $\alpha \in \{0.1, 0.05\}$ with 20 clients in Tables 1, 3, and 4. We visualize the data distributions in Figure \ref{fig:data_distr}. It shows that in both settings, the clients have extremely heterogeneous data distributions. Especially when $\alpha=0.05$, all clients have some classes missing, and some clients have extremely rare data (e.g. client 1). We note that these settings are very realistic in practical FL scenarios.

\end{document}